\documentclass[3p, 12pt, onecolumn, authoryear]{elsarticle}

\usepackage{todonotes}
\usepackage{changepage}
\usepackage{amsmath}
\usepackage{pdflscape}
\usepackage{amssymb}
\usepackage{pifont}
\usepackage{tabularx}
\usepackage{titlesec}
\usepackage{caption}
\usepackage{subcaption}

\usepackage{xcolor}
\usepackage{mathtools}
\usepackage[hyphens]{url}
\usepackage{ragged2e}
\usepackage{pgfplotstable}
\usepackage{booktabs}
\usepackage{longtable}
\usepackage{subcaption}
\usepackage{chngcntr}
\usepackage{hyperref}
\usepackage{makecell}
\usepackage{multirow}
\usepackage{multicol}
\usepackage[figuresright]{rotating}
\usepackage{pgfplotstable}
\usepackage{colortbl}
\usepackage{arydshln}
\hypersetup{
	colorlinks=true,
	linkcolor=blue,
	filecolor=blue,      
	urlcolor=blue,
	citecolor=blue,
	breaklinks=true
}

\journal{Data Mining and Knowledge Discovery}

\newcommand{\cmark}{\textcolor{green}{\ding{51}}}%
\newcommand{\xmark}{\textcolor{red}{\ding{55}}}%
\newcommand{\DAG}{\textsuperscript{\textdagger}}%

\titleformat{\paragraph}
{\normalfont\normalsize\itshape}{\theparagraph}{1em}{}
\titlespacing*{\paragraph}
{0pt}{3.25ex plus 1ex minus .2ex}{1.5ex plus .2ex}

\begin{document}
\begin{frontmatter}
		
\title{Forecast Evaluation for Data Scientists: Common Pitfalls and Best Practices}

\author[rmit]{Hansika Hewamalage}
\author[business]{Klaus Ackermann}
\author[it]{Christoph Bergmeir\corref{cor1}}
\address{hansika.hewamalage@rmit.edu.au, Klaus.Ackermann@monash.edu, Christoph.Bergmeir@monash.edu }
\address[rmit]{School of Computing Technologies, RMIT University, Melbourne, Australia}
\address[business]{SoDa Labs and Dept of Econometrics \& Business Statistics, Monash University, Melbourne, Australia.}
\address[it]{Dept of Data Science and AI, Faculty of IT, Monash University, Melbourne, Australia.}

\cortext[cor1]{Corresponding Author Name: Christoph Bergmeir, Affiliation: Dept of Data Science and AI, Faculty of IT, Monash University, Melbourne, Australia, Postal Address: 
Faculty of Information Technology, Monash University, 20 Exhibition Walk, Clayton Campus, Wellington Road, Clayton VIC 3800, Australia, E-mail address: christoph.bergmeir@monash.edu}

\begin{abstract}
	Machine Learning (ML) and in particular Deep Learning (DL) methods nowadays are increasingly replacing traditional methods in many different domains involved with important decision making activities. Sophisticated DL techniques tailor-made for specific tasks such as image recognition, signal processing, or speech analysis are being introduced at a fast pace with many improvements. However, for the domain of time series forecasting, the current state in the ML community is perhaps where other domains such as Natural Language Processing and Computer Vision were at several years ago. The field of forecasting has mainly been fostered by statisticians/econometricians; consequently the related concepts are not the mainstream knowledge among general ML practitioners. The different forms of non-stationarities associated with time series challenge the capabilities of data-driven ML models. 
	 Nevertheless, recent trends in the domain have demonstrated that with the availability of massive amounts of time series, ML and DL techniques are quite competent in time series forecasting, when related pitfalls are properly handled. Therefore, in this work we provide a tutorial-like compilation of the details of one of the most important steps in the overall forecasting process, namely the evaluation. This way, we intend to impart the information associated with forecast evaluation to fit the context of ML, as means of bridging the knowledge gap between traditional methods of forecasting and current state-of-the-art ML techniques. We elaborate the details of the different problematic characteristics of time series such as non-normalities and non-stationarities and how they are associated with common pitfalls in forecast evaluation. Best practices in forecast evaluation are outlined with respect to the different steps such as data partitioning, error calculation, statistical testing, and others. Further guidelines are also provided along selecting valid and suitable error measures depending on the specific characteristics of the dataset at hand.
	
\end{abstract}
\end{frontmatter}




\section{Introduction}
In the present era of Big Data, Machine Learning (ML) and Deep Learning (DL) based techniques are driving the automatic decision making in many domains such as Natural Language Processing (NLP) or Time Series Classification \citep[TSC, ][]{Bagnall2016, IsmailFawaz2019}. Although fields such as NLP and Computer Vision have heavily been dominated by ML and DL based techniques for decades by now, this has hardly been the case for the field of forecasting, until very recently. Forecasting was traditionally the field of statisticians and econometricians.
However, nowadays, with many companies hiring data scientists, often these data scientists are tasked with forecasting. Therefore, now in many situations practitioners are tasked with forecasting that have a good background in ML and data science, but that are not aware of the decades of research in the forecasting space.
This involves many aspects of the process of forecasting, from the point of data pre-processing, building models to final forecast evaluation. Due to the self-supervised and sequential nature of forecasting tasks, it is often associated with many pitfalls that usual ML practitioners are not aware of. Out of all these aspects, in this particular work, we focus on the evaluation of point forecasts as a key step in the overall process of forecasting.


Evaluating the performance of models is key to the development of concepts and practices in any domain. The general process involves employing a number of models having different characteristics, training them on a training dataset and then applying them on a validation set afterwards. Then, model selection may be performed by evaluating on the validation set to select the best models. Otherwise, ensemble models may be developed instead, by combining the forecasts from all the different models, and usually a final evaluation is then performed on a test set. 
In research areas such as classification and regression, there are well-established standard practices for evaluation. Data partitioning is performed by using a standard k-fold Cross-Validation (CV) to tune the model hyperparameters based on the error on a validation sets, the model with the best hyperparameter combination is tested on the testing set, standard error measures such as squared errors, absolute errors or precision, recall, area under curve are computed and finally the best models are selected. 
These best methods may continue to deliver reasonable predictions for a certain problem task, i.e., they generalise well, under the assumption that there are no changes of the distribution of the underlying data, which otherwise would need to be addressed as concept drift~\citep{webb_concept_drift, Ghomeshi2019, Ikonomovska2010} or non-stationarity.


In contrast, evaluating forecasting models can be a surprisingly complicated task. 
Data partitioning and model selection have many different options in the context of forecasting, including fixed origin, rolling origin evaluation and other CV setups as well as controversial arguments associated with them. 
Due to the inherent non-independence, non-stationarities and non-normalities of time series, these choices are complex. Also, most error measures are susceptible to break down under certain of these conditions.
Other considerations are whether to summarise errors across all available time series or consider different steps of the forecast horizon separately etc. 
As a consequence, without wanting to call them out here, we regularly come across papers in top AI/ML conferences and journals (even winning best paper awards) that use inadequate and miss-leading benchmark methods for comparison (e.g., non-seasonal models for long-term forecasting on seasonal series), others that use mean absolute percentage error (MAPE) for evaluation with series, e.g., with values in the $[-1,\ 1]$ interval because the authors think the MAPE is a somewhat generic ``time series error measure'', even though MAPE is clearly inadequate in such settings. Other works make statements along the lines of ARIMA being able to tackle non-stationarity whereas ML models can't, neglecting that the only thing ARIMA does is a differencing of the series as a pre-processing step to address non-stationarity. A step that can easily be done as preprocessing for any ML method as well.
In other works, we see methods compared using MAE as the error measure, and only the proposed method by those authors is trained with L1 loss, all other competitors with L2 loss, which leads to unfair comparisons as the L1 loss optimises towards MAE, whereas the L2 loss optimises towards RMSE. Many other works evaluate on a handfull of somewhat randomly picked time series and then show plots of forecasts versus actuals as ``proof'' of how well their method works, without considering simple benchmarks or meaningful error measures, and other similar problems.
Also, frequently forecasting competitions and research works introduce new evaluation measures and methodologies, sometimes neglecting the prior research, e.g., by seemingly not understanding that dividing a series by its mean will not solve scaling issues for many types of non-stationarities (e.g., strong trends). 
Thus, there is no generally accepted standard for forecast evaluation in every possible scenario. This gap has harmed progress in ML methods for forecasting significantly in the past. It is damaging the area currently, with spurious results in many papers, with researchers new to the field not being able to distinguish between methods that work and methods that don't, and the associated slower progress and waste of resources.


Overall, this article makes an effort in the direction of raising awareness among ML practitioners regarding the best practices and pitfalls associated with the different steps of the point forecast evaluation process. Similar exhaustive efforts have been taken in the literature to review, formally define and categorise other important concepts in the ML domain such as concept drift~\citep{webb_concept_drift} and mining statistically sound patterns from data~\citep{Hmlinen2018}. The rest of this paper is structured as follows. Section \ref{sec:terminology} first introduces terminology associated with the domain of forecasting. Next, Section \ref{sec:motivation} details the motivation for this article, including an introduction to the different forms of non-stationarities/non-normalities seen in time series data, along with common pitfalls related to using competitive benchmarks, visualisation of results using forecast plots and avoiding data leakage in forecast evaluation. Then, Section \ref{sec:overview} presents an overview of the process of forecast evaluation. In Section \ref{sec:evaluation_setup}, we provide a tutorial/guideline around how to best partition the data for a given forecasting problem such that it is not affecting the sequential nature or the non-stationarities involved with the problem. Section \ref{sec:error_measures} first presents a comprehensive literature review of many different evaluation measures proposed over the years. This is supplemented by a critical analysis of how each of them can break/fail under different circumstances of the time series. This section also provides a guideline on selecting evaluation measures depending on the characteristics of the time series under consideration. In Section \ref{sec:statistical_tests}, we provide details of popular techniques used for statistical testing for significance of differences between models. Finally, Section \ref{sec:conclusions} concludes the paper by summarising the overall content of the paper and highlighting the best practices for forecast evaluation.

\section{Problem Definition and Terminology}
\label{sec:terminology}

The scope of the discussion in this article focusses on point forecasting, where the interest is to predict one particular statistic (mean/median) of the overall forecast distribution. However, we note that there are many works in the literature around predicting distributions and evaluating accordingly. In this section we provide a general overview of the terminology used in the context of time series forecasting.

Throughout this article, we focus on the task of \textit{univariate forecasting}. Univariate forecasting is when future values of a time series are predicted using the past values of that same series as well as some other exogenous time varying variables which may affect the target series. This can be formulated as in Equation \ref{eqn:univariate_forecasting}.

\begin{equation}
	\label{eqn:univariate_forecasting}
	\hat{y}_{t+h} = g(\mathcal{X}_t, \theta)
\end{equation}

\begin{figure*}[htbp!]
	\centering
	\includegraphics[scale=1.5]{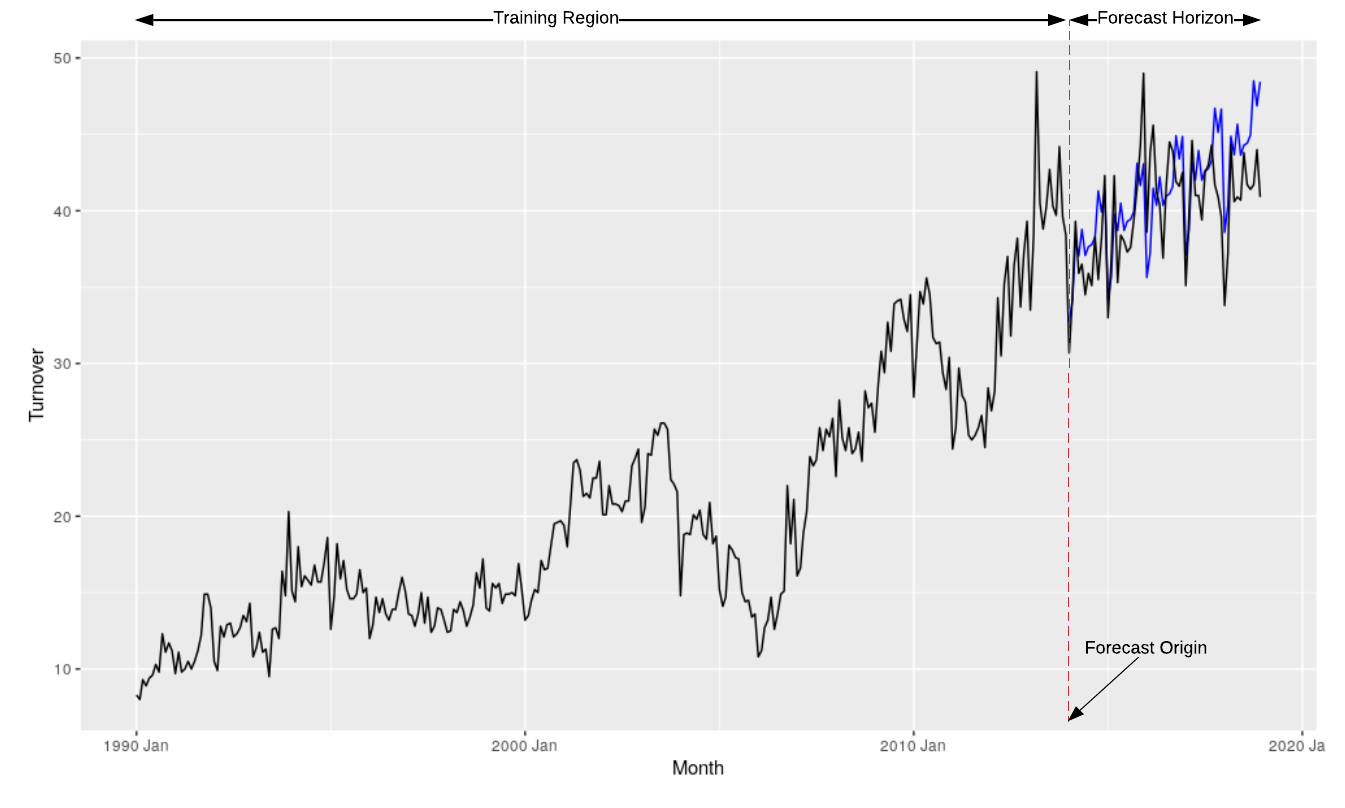}
	\caption{A Forecasting Scenario with Training Region of the Data, Forecast Origin and the Forecast Horizon}
	\label{fig:forecasting_intro}
\end{figure*}

Here, $g$ is a (non-linear, non-parametric) function, for example an ML model and $\theta$ are its parameters. $\mathcal{X}_t$ are all input data and information available to the model up until time $t$ where $t$ is the \textit{forecast origin}. Therefore, forecast origin is the last known data point from which the forecasting begins. $h$ denotes the \textit{forecast horizon}, i.e.\ the length of the time period into the future for which forecasting of the target value is performed. These are indicated in Figure \ref{fig:forecasting_intro}.

Traditionally, in univariate forecasting without external variables, $\mathcal{X}_t={\bf y}=y_1,\ldots, y_t$. With external variables, $\mathcal{X}_t$ contains also the values of the external variables, up to time $t$ or also future values if known. We do not consider multivariate regression in this work, where to predict the future values of many target series together, past values of all series as well as other potentially available external variables are used.

\begin{figure*}[htbp!]
	\centering
	\includegraphics[scale=1.4]{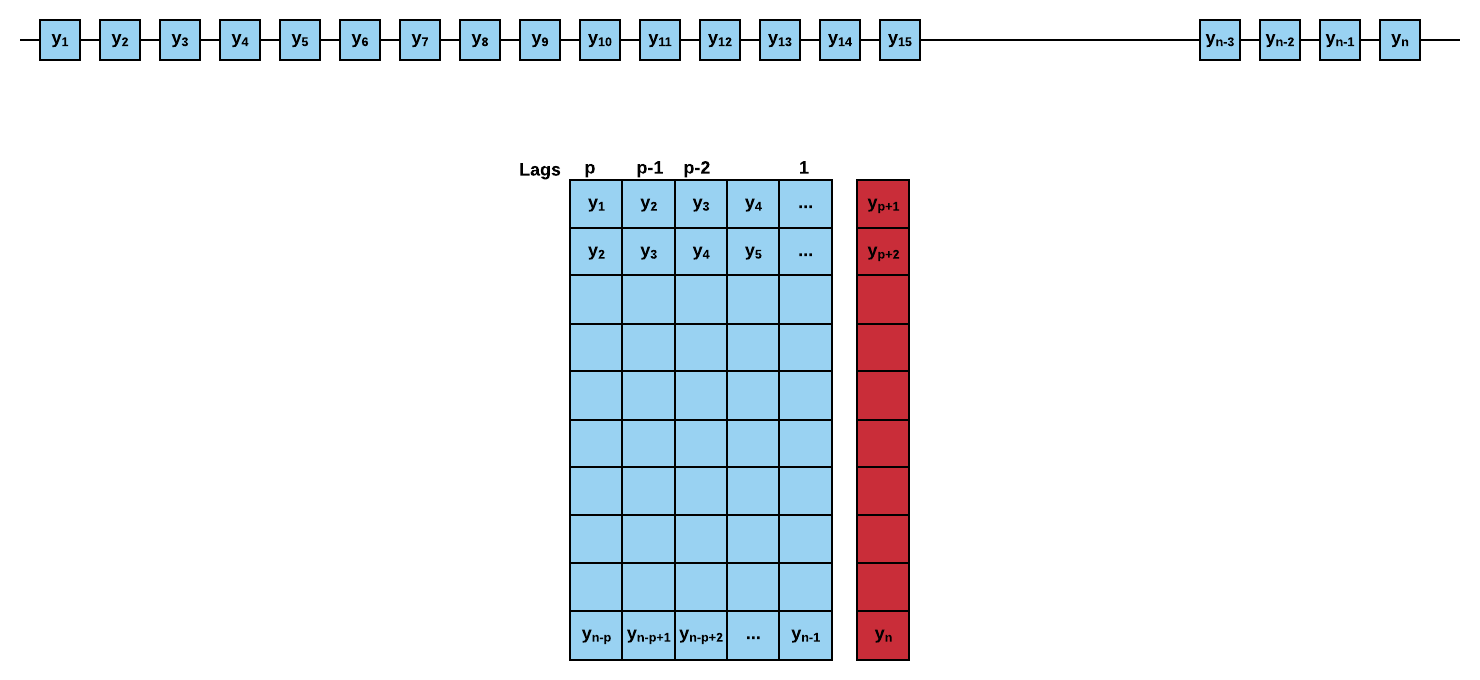}
	\caption{Embedded Matrix for AR Process of Order $p$}
	\label{fig:embedded_matrix}
\end{figure*}

In a time series context, we define a \textit{lag} with respect to a time step $t$ as the values of the series at previous time steps. For example, \textit{lag 1} is the value at time step $t-1$ and \textit{lag m} is the value at time step $t-m$. In a so-called \textit{Auto-Regression (AR)}, $\mathcal{X}_t$ in Equation \ref{eqn:univariate_forecasting} only goes back a fixed amount of lags, usually called the \textit{model order}. The very name indicates that the regression is performed against the values of the target series itself. An auto-regression of a time series uses an \textit{embedded matrix}. In the embedded matrix in Figure \ref{fig:embedded_matrix}, when the model order is $p$, every row has $p+1$ consecutive observations from the time series. During model training, every row is considered a separate data instance, where values at lags $1, 2,...\ p$ are considered predictors for the target quantity of the time series at time step $p+1$. The process goes through the whole series, shifting the target quantity by one time step in each row, to form a matrix. Therefore, in an AR setup of order $p$ on a series of length $n$, the number of data instances for training is equivalent to $n-p$.

\begin{figure*}[htbp!]
	\centering
	\includegraphics[scale=1.3]{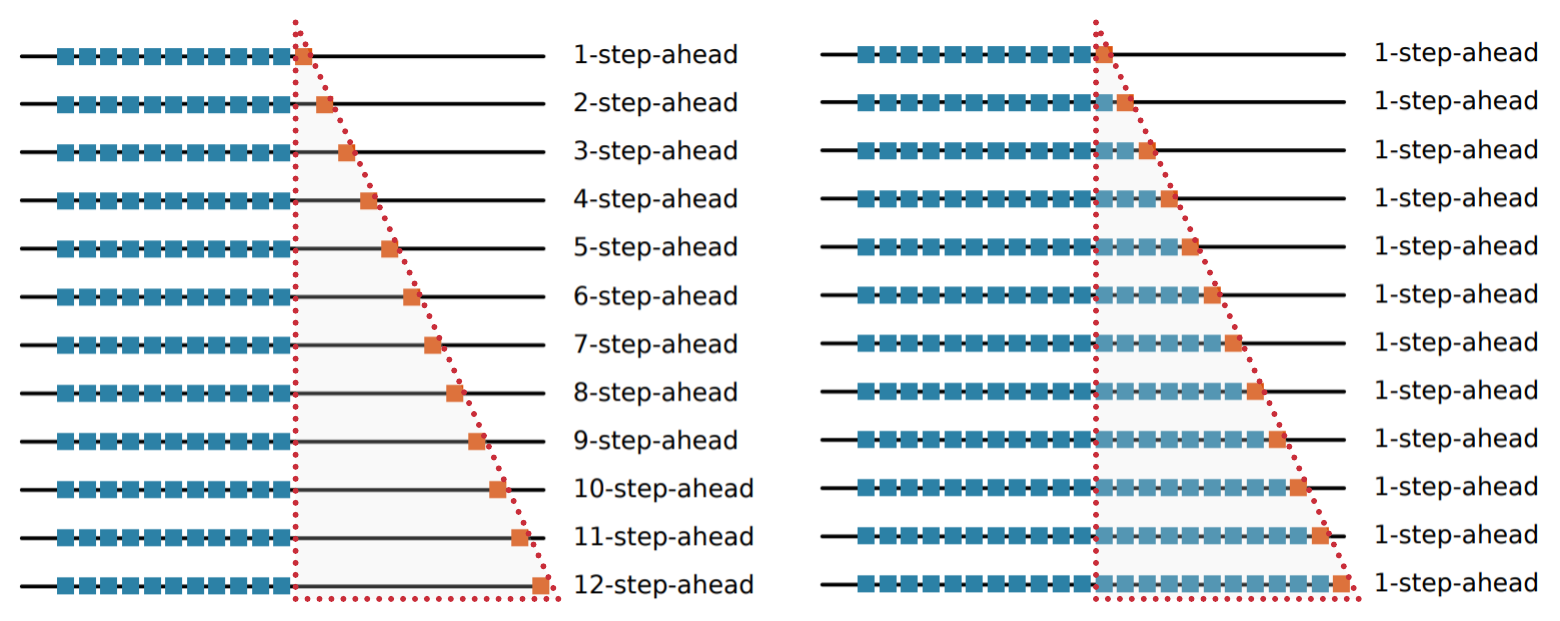}
	\caption{Comparison of fixed origin vs. rolling origin setups. The blue and orange data points represent the training and testing sets respectively at each evaluation. The figure on the left side shows the fixed origin setup where the forecast origin remains constant. The figure on the right shows the rolling origin setup where the forecast origin rolls forward and the forecast horizon is constant. The red dotted lined triangle encloses all the time steps used for testing across all the evaluations. Compared to the fixed origin setup, it is seen that in the rolling origin setup, testing data instances in each evaluation pass on to the training set in the next evaluation step. }
	\label{fig:rolling_vs_fixed}
\end{figure*}

Using this notion of univariate forecasting, both \textit{local models} and \textit{global models} can be developed. For a local model, the parameters $\theta$ of the model are trained only using one series. Therefore, in the scenario where many time series are available, local models need to be developed as one per each series. On the other hand, for global models, the parameters $\theta$ are trained across time series, i.e., using data from all the series. Thus, the embedded matrix contains data instances from many series. However, for the prediction of a single series, $\mathcal{X}_t$ in Equation \ref{eqn:univariate_forecasting} only considers the corresponding lags of that particular series~\citep{Januschowski2020ijf}. Similar to other ML tasks, validation and test sets are used for hyperparameter tuning of the models and for testing. Evaluations on validation and test sets are often called \textit{out-of-sample (OOS)} evaluations in forecasting. The two main setups for OOS evaluation in forecasting are \textit{fixed origin evaluation} and \textit{rolling origin evaluation}~\citep{TASHMAN2000437}. 
Figure \ref{fig:rolling_vs_fixed} shows the difference between the two setups. In the fixed origin setup, the forecast origin is fixed as well as the training region, and the forecasts are computed as one-step ahead or multi-step ahead depending on the requirements. In the rolling origin setup, the size of the forecast horizon is fixed, but the forecast origin changes over the time series (rolling origin), thus effectively creating multiple test periods for evaluation. With every new forecast origin, new data becomes available for the model which can be used for re-fitting of the model. However, as seen on Figure \ref{fig:rolling_vs_fixed}, since the rolling origin setup allows the data to pass on from the testing set to the training set of the next consecutive evaluation step, this setup, if not used properly, is naturally susceptible to data leakage dangers, where information of the future may leak to the model training phase using past data (further discussed in Section \ref{sec:leakage}). The rolling origin setup is also called \textit{time series cross-validation (tsCV)} and \textit{prequential evaluation} in the literature \citep{robgeorg2018otext,gama2013evaluating}. Further details of these approaches are discussed in Section \ref{sec:evaluation_setup}. 


\section{Motivation and Common Pitfalls}
\label{sec:motivation}

This section is devoted to provide the motivation of our work and we discuss the general problems faced in forecasting, in comparison to a usual ML task.

\subsection{Characteristics of Time Series}
\label{sec:characteristics_of_timeseries}

What makes time series forecasting a more difficult problem in comparison to other ML tasks, are the different \textit{non-stationarities} and \textit{non-normalities} commonly embedded in time series. Listed below are some of such possibly problematic characteristics of time series. 

\begin{enumerate}
	\item Non-stationarities.
	\begin{itemize}
		\item Seasonality
		\item Trends (Deterministic, e.g., Linear/Exponential)
		\item Stochastic Trends / Unit Roots
		\item Heteroscedasticity
		\item Structural Breaks (sudden changes, often with level shifts)
	\end{itemize}
	\item Non-normality
	\begin{itemize}
		\item Non-symmetric distributions
		\item Fat tails 
		\item Intermittency
		\item Outliers
	\end{itemize}
	\item Series with very short history
	
\end{enumerate}
Non-stationarity in general means that the distribution of the data in the time series is not constant, but it changes depending on the time \citep[see, e.g., ][]{Salles2019-ql}. What we refer to as non-stationarity in this work is the violation of strong stationarity defined as in Equation \ref{eqn:strong_stationarity}~\citep{Cox2017}. Strong stationarity is defined as the distribution of a finite window (sub-sequence) of a time series (discrete-time stochastic process) remaining the same as we shift the window across time. In Equation \ref{eqn:strong_stationarity}, $y_{t}$ refers to the time series value at time step $t$; $\tau \in \mathbb{Z}$ is the size of the shift of the window and $n \in \mathbb{N}$ is the size of the window. $F_Y(y_{t+\tau}, y_{t+1+\tau}, ..., y_{t+n+\tau})$ refers to the cumulative distribution function of the joint distribution of $(y_{t+\tau}, y_{t+1+\tau}, ..., y_{t+n+\tau})$. Hence, according to Equation \ref{eqn:strong_stationarity}, $F_Y$ is not a function of time, it does not depend on the shift of the window. In the rest of this paper, we refer to the violation of strong stationarity simply as non-stationarity.

\begin{equation}
	\label{eqn:strong_stationarity}
	F_Y(y_{t+\tau}, y_{t+1+\tau}, ..., y_{t+n+\tau}) = F_Y(y_{t}, y_{t+1}, ..., y_{t+n}), \text{		for all $\tau \in \mathbb{Z}$ and $n \in \mathbb{N}$}
\end{equation}

\begin{figure*}[htb]
	\hspace{-0.5cm}
	\includegraphics[width=\textwidth]{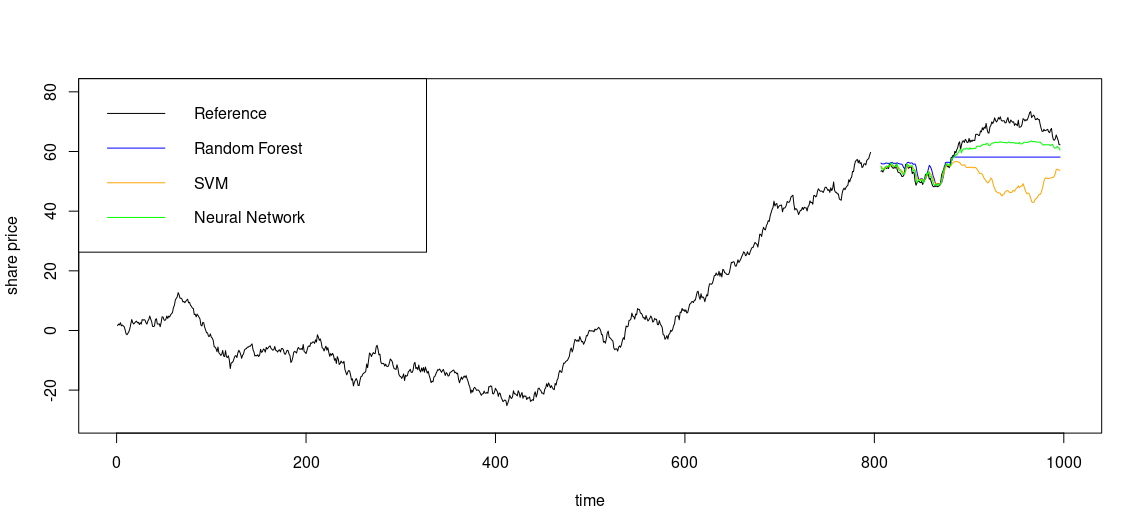}
	\caption{Forecasts from different models on a series with unit root based non-stationarity, with stochastic trends. In this example, we have a continuously increasing series (increasing mean) due to the unit root. The ML models are built as autoregressive models without any pre- or post-processing, and as such have very limited capacity to predict values beyond the domain of the training set.  
}
	\label{fig:unitroot_example}
\end{figure*}

\begin{figure*}[htb]
		\hspace{-1cm}
	\includegraphics[width=\textwidth]{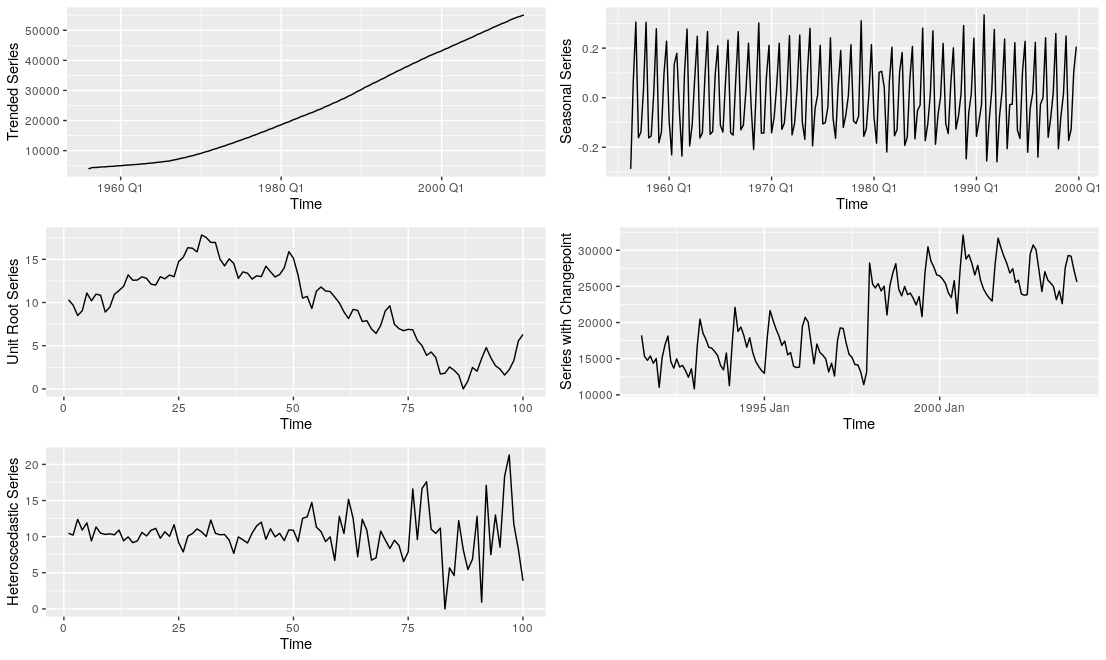}
	\caption{Different Non-stationarities of Series}
	\label{fig:non_stationarities}
\end{figure*}

Figure~\ref{fig:unitroot_example} gives an example of possible problems when building ML models on such data, where the models fail to produce reasonable forecasts as the range of values is different in the training and test sets. Different types of non-stationarities are illustrated in Figure \ref{fig:non_stationarities}. \textit{Seasonality} usually means that the mean of the series changes periodically over time, with a fixed length periodicity. Trends can be twofold; 1) \textit{deterministic trends} - change the mean of the series 2) \textit{stochastic trends} (resulting from unit roots) - change both the mean and variance of the series~\citep{Salles2019-ql}. Note that neither trend nor seasonality are concepts that have precise formal definitions. They are usually merely defined as smoothed versions of the time series, where for the seasonality the smoothing occurs over particular seasons (e.g., in a daily series, the series of all Mondays needs to be smooth, etc.). \textit{Heteroscedasticity} changes the variance of the series and \textit{structural breaks} can change the mean or other properties of the series. \textit{Structural break} is a term used in Econometrics and Statistics in a time series context to describe a sudden change in the series. It therewith has considerable overlap with the notion of \textit{sudden concept drift} in an ML environment, where a sudden change of the data distribution is observed \citep{webb_concept_drift}. 

On the other hand, data can be far from normality, for example having fat tails, or when conditions such as outliers or intermittency are observed in the series. Non-stationarities and non-normalities are both seen quite commonly in many real-world time series and the decisions taken during forecast evaluation depend on which of these characteristics the series have. There is no single universal rule that applies to every scenario.

\subsection{Benchmarks for Forecast Evaluation}
\label{sec:benchmarks}

\begin{figure*}[htbp!]
	\centering
	\begin{subfigure}{\textwidth}
		\centering
		\includegraphics[scale=0.5]{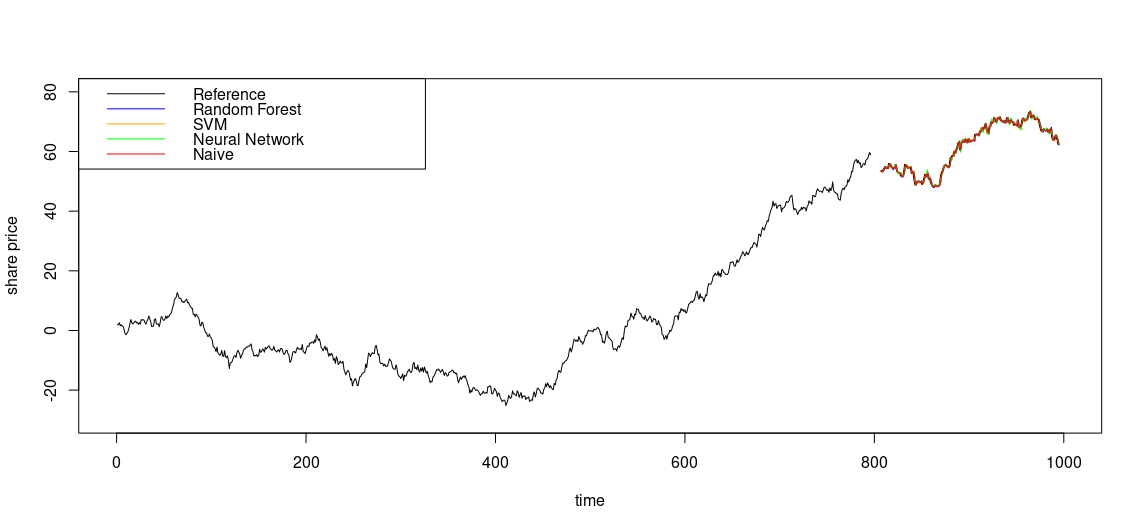}
		\caption{Series with unit root based non-stationarity and forecasts from different models}
		\label{fig:unitroot_example2}
	\end{subfigure}
	\hfill
	\begin{subfigure}{\textwidth}
		\centering
		\includegraphics[scale=0.5]{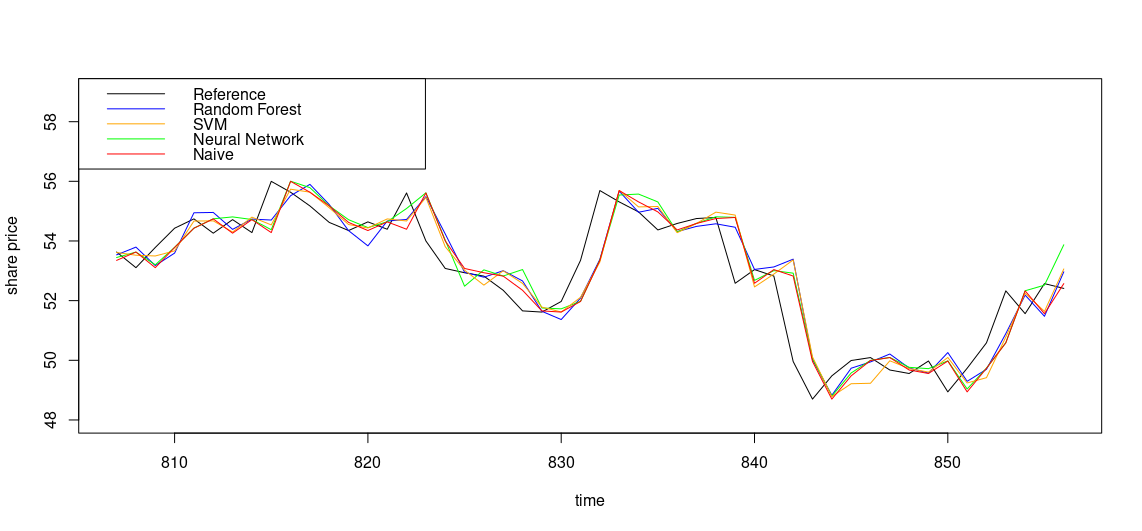}
		\caption{Forecasts from the different models on a selected subset of the test set (timestamps 807 - 856)}
		\label{fig:unitroot_example2_test_set}
	\end{subfigure}
	\caption{Forecasts from different models on a series with unit root based non-stationarity, with stochastic trends. The ML models are built as autoregressive integrated models, i.e., differencing has been done as pre-processing. The methods show very similar behaviour to the na\"ive forecast, and do not add any value over it by definition of the Data Generating Process used.
		\label{fig:unitroot_example_with_difference} 
}
\end{figure*}

Benchmarks are an important part of forecast evaluation. Comparison against the right benchmarks and especially the simpler ones is essential. Arguably the simplest benchmark that is commonly employed in forecasting is the na\"ive forecast, also called persistence model or no-change model, that simply uses the last known observation as the forecast. It has demonstrated competitive performance in many scenarios~\citep{Scott_Armstrong_undated}. Figure \ref{fig:unitroot_example_with_difference} illustrates the behaviour of different models that have been trained with differencing as appropriate preprocessing on a series that has a unit root based non-stationarity. If the series has no further predictable properties above the unit root, i.e., it is a random walk where the innovation added to the last observation follows a normal distribution with a mean of zero, the na\"ive forecast is the theoretically best forecast. Other, more complex forecasting methods in this scenario will have no true predictive power beyond the na\"ive method, and any superiority, e.g., in error evaluations is by pure chance, and should be able to be identified as a spurious result on sufficiently large datasets.

Equation \ref{eqn:random_walk} shows the definition of a random walk, where $\epsilon_t$ is white noise; i.e.\ sampled from a normal distribution. Accordingly, the na\"ive forecast at any time step in the horizon can be defined as in Equation \ref{eqn:naive_forecast}.
As the na\"ive forecast is the last known observation, the forecast is a shifted version of the time series where the forecast simply follows the actuals (see Figure~\ref{fig:unitroot_example2_test_set}).

\begin{equation}
	\label{eqn:random_walk}
	y_{t+1} = y_{t} + \epsilon_t
\end{equation}

\begin{equation}
\label{eqn:naive_forecast}
\hat{y}_{t+h} = y_t
\end{equation}

In many practical applications, we find series that show strongly integrated behaviour and therewith are close to random walks (such as stock market data, wind power, wind speed). Here, a na\"ive forecast is a trivial yet competitive benchmark and without comparing against it, quality of more complex models cannot be meaningfully assessed. Furthermore, also more complex methods will in such series usually show a behaviour where they mostly follow the series in the same way as the na\"ive forecast, and improvements are often small percentages over the performance of the na\"ive benchmark.
 
As such, the benchmarks and the error measure used play an important role in such a setting. For instance, by using a relative error measure (detailed further in Section \ref{sec:error_measures}) that lets us directly compare against a simple benchmark such as the na\"ive, we can be certain of the competitiveness of the model against simple methods. On series that have clear seasonal patterns, models should accordingly be benchmarked against the seasonal na\"ive model as the most simplistic benchmark, and also other simple benchmarks are commonly used in forecasting. 



%
%

\subsection{Forecast Plots}

Plots with time series forecasting results can be quite misleading and should be used with caution. Analysing plots of forecasts from different models along with the actuals and concluding that they seem to fit well can lead to wrong conclusions. It is important to use benchmarks and evaluation metrics that are right for the context. Even with good error measures, in a scenario like a random walk series as in Figure \ref{fig:unitroot_example_with_difference}, as stated before, our models may achieve better accuracy than the na\"ive method, but it will be a spurious result.

The visual appeal of a generated forecast or the possibility of such a forecast to happen in general are not good criteria to judge forecasts. 

\begin{figure}[htb]
     \centering
     \begin{subfigure}[b]{0.7\textwidth}
         \centering
         \includegraphics[width=\textwidth]{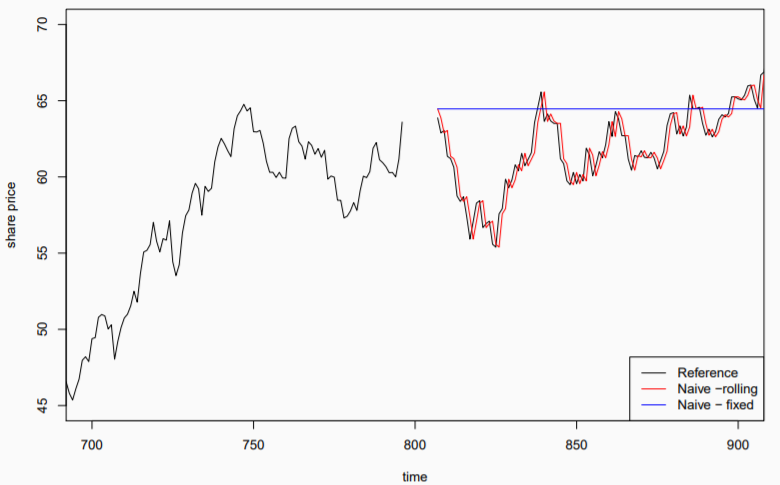}
         \caption{Rolling Origin vs. Fixed Origin Comparison for the Na\"ive Forecast}
         \label{fig:naive_rolling_fixed}
     \end{subfigure}
     \hfill
     \begin{subfigure}[b]{0.62\textwidth}
         \centering
         \includegraphics[width=\textwidth]{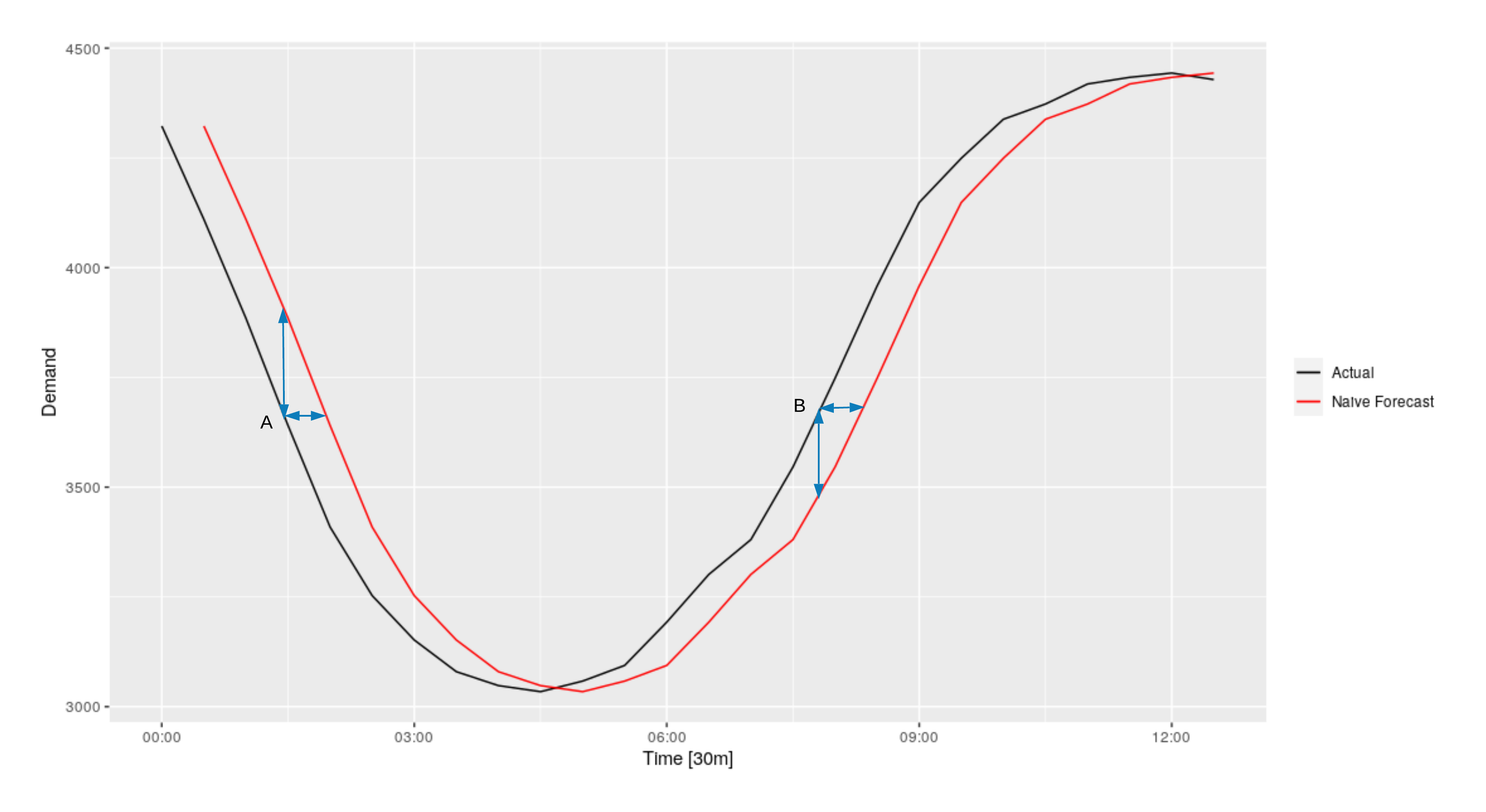}
         \caption{Visual Delusion of the Na\"ive Forecast}
         \label{fig:naive_plot_issues}
     \end{subfigure}
     \hfill
     \begin{subfigure}[b]{0.36\textwidth}
         \centering
         \includegraphics[width=\textwidth]{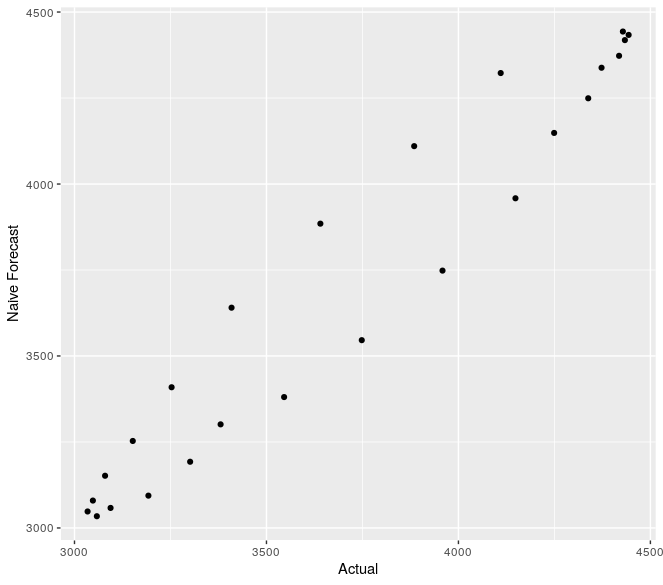}
         \caption{Scatter plot of Actuals against the Na\"ive Forecast}
         \label{fig:scatterplot}
     \end{subfigure}
        \caption{Properties of the na\"ive forecast}
\end{figure}

Figure \ref{fig:naive_rolling_fixed} shows another random walk series, with the na\"ive forecast as the best forecast by definition of the Data Generating Process (DGP). The figure furthermore shows the forecasts under fixed origin and rolling origin data partitioning schemes.
When periodic re-fitting is done with new data coming in as in a rolling origin setup, the na\"ive forecast gets continuously updated with the last observed value. For the fixed origin context on the other hand, the na\"ive forecast remains constant as a straight line corresponding to the last seen observation in the training series. 
We see that with a rolling-origin na\"ive forecast, the predictions tend to look visually very appealing, as the forecasts follow the actuals and our eyes are deceived by the smaller horizontal distances instead of the vertical distances that are relevant for evaluation. Figure \ref{fig:naive_plot_issues} illustrates this behaviour. It is clear how the horizontal distance between the actuals and the na\"ive forecast at both points A and B are much less compared to the vertical distances which are the relevant ones for evaluation. If a scatter-plot of the actuals against the forecasts as in Figure \ref{fig:scatterplot} can be used instead, it may give a much better picture of where the forecasts stand with respect to reality, by discarding the time domain. On the other hand, on a series with integrated behaviour, the na\"ive method is a strong benchmark and other competitive methods on such a series will also tend to show behaviours of following the actuals. 
%
%
In these situations we need to rely on the error measures, as the plots do not give us much information.


Figure~\ref{fig:naive_rolling_fixed} shows another issue with forecasts, as the na\"ive forecast for fixed origin is a constant. Although this does not look realistic, and in most application domains we can be certain that the actuals will not be constant, practitioners may mistakenly identify such behaviour as a potential problem with the models, where this forecast is indeed the best possible forecast in the sense that it minimises the error based on the information available at present.

%
%
%
%

In summary, plots of the forecasts can be deceiving and should be used mostly for sanity checking. Decisions should mostly be made based on evaluations with error measures and not based on plots.


\subsection{Data Leakage in Forecast Evaluation}
\label{sec:leakage}

Data leakage refers to the inadvertent use of data from the test set, or more generally data not available during inference, while training a model. It is always a potential problem in any ML task. For example, \citet{KaufmanShachar2012Lidm} present an extensive review on the concept of data leakage for data mining and potential ways to avoid it. \citet{Arnottjfds.2019.1.064} discuss this in relation to the domain of finance.  \citet{pmlr-v161-hannun21a} propose a technique based on Fisher information that can be used to detect data leakage of a model with respect to various subsets of the dataset. \citet{brownlee2020data} also provide a tutorial overview on data preparation for common ML applications while avoiding data leakage in the process. However, in forecasting data leakage can happen easier and can be harder to avoid than in other ML tasks such as classification/regression. 

Forecasting is usually performed in a self-supervised manner with rolling origin evaluations where periodic re-training of models is performed, and within this re-training, it is normal that data travels from the test to the training set. As such, it is often difficult and not practical to separate training and evaluation code bases. As such, we often have to trust the software provider that everything is implemented correctly, and an external evaluation is difficult.

Also, more indirect forms of data leakage can happen in forecasting. In analogy to classification/regression, where data leakage sometimes happens by normalising data before partitioning for cross-validation, in forecasting, data leakage can happen by performing smoothing, decomposition (mode decomposition), normalisation etc.\ over the whole series before partitioning for training and testing. Data leakage can happen even when extracting features such as \texttt{tsfeatures}~\citep{tsfeatures}, \texttt{catch22}~\citep{Lubba2019} that are not constant over time, to feed as inputs to the model. 
Thus, features can be extracted only from the training set data, and may need to be re-calculated either periodically or over the specific input windows. However, this can be computationally expensive. 

Another type of leakage especially when training global models that learn across series, which is common practice nowadays for ML models, is when one series in the dataset contains information about the future of another series. For example with an external shock like COVID-19 or a global economy collapse, all the series in the dataset can be equally affected. Therefore, if the series in the dataset are not aligned and one series contains the future values with respect to another, when splitting the training region, future information can be already included within the training set. However, in real world application series are usually aligned so that this is not a big problem. On the other hand, in a competition setup such as the M3 and M4 forecasting competitions~\citep{MAKRIDAKIS2000451, makridakis2020m4}, where the series are not aligned, this can easily happen. 

Data leakage can also happen simply due to using the wrong forecast horizon. This can happen by using data that in practice will become available later. For example, we could build a one-day-ahead model, but use summary statistics over the whole day. This means that we cannot run the model until midnight, when we have all data from that day available. If the relevant people who use the forecasts work only from 9am-5pm, it becomes effectively a same-day model. The other option is to set the day to start and end at 5pm everyday, but that may lead to other problems.

In conclusion, data leakage dangers are common in self-supervised forecasting tasks. It is important to avoid leakage problems 1) in rolling origin schemes by being able to verify and trust the implementation, as external evaluation can be difficult 2) during preprocessing of the data (normalising, smoothing etc.) and extracting features such as \texttt{tsfeatures} by splitting the data into training and test sets beforehand 3) by making sure that within a set of series, one series does not contain in its training period potential information about the future of another series.

\section{Overview of the Forecast Evaluation Process}
\label{sec:overview}

Forecast model building and evaluation typically encompasses the following steps. 

\begin{itemize}
	\item Data partitioning
	\item Forecasting
	\item Error Calculation
	\item Error Measure Calculation
	\item Statistical Tests for Significance (optional)
	\item Model Selection (optional)
\end{itemize}


The process of evaluation in a usual regression problem is quite straightforward. The models fitted to the training dataset output a prediction for a single target value in the validation set, an error such as the quadratic loss is computed for each prediction and target value combination, and finally the errors from all the predictions in the validation set are summarised using some error measure such as the Root Mean Squared Error (RMSE). The best model out of the pool of fitted models is selected based on the value of this final error measure on the validation set. The relevant error measures used etc.\ are standard and established as best practices in these domains.

However, when it comes to forecast evaluation, many different options are available for each of the  aforementioned steps and no standards have been established thus far, although several pitfalls associated with certain evaluation setups have been identified. Two other optional activities related to forecast evaluation are Statistical Tests for Significance and Model Selection. They are not performed always by practitioners. Instead of selecting one best model, we may sometimes be interested in deploying an ensemble of all the models. Out of these different steps in evaluation, in this article we discuss the approaches commonly used for Data Partitioning, Model Selection, Error and Error Measure Calculation as well as Statistical Tests for Significance.

\section{Data Partitioning}
\label{sec:evaluation_setup}

When performing OOS evaluation in forecasting (using a validation or test set), we can either evaluate for every individual forecast step separately (one-step-ahead error, two-step-ahead error) or the whole test period on average depending on the forecasting scheme of the underlying models.  
In this section, we explain the details of different data partitioning techniques for evaluations performed in the context of forecasting. We also explain the options for model selection based on these different data partitioning strategies.

\subsection{Fixed Origin Setup}
Fixed origin setup is a faster and easier to implement evaluation setup. Fixed origin setup is the usual setup used in many academic contexts such as forecasting competitions since this can effectively avoid data leakage problems as the test set is not disclosed in any way. In fact for competition scenarios where the dates of the series are not aligned (like the M3, M4 competitions), a fixed origin setup may be sufficient. However, for many practical scenarios, a fixed origin setup is problematic and may not be a sufficient evaluation. With a single series, the fixed origin setup only provides one forecast per each forecast step in the horizon. According to \citet{TASHMAN2000437}, a preferred characteristic of OOS forecast evaluation is to have sufficient forecasts at each forecast step.
Furthermore, for a single series, the testing period is short unless a long forecast horizon is used. However, with a long forecast horizon, the problem is that we are mixing very different forecasts. For example, for a 400-step-ahead forecast, short-term dynamics due to autocorrelation may be irrelevant, and trend and seasonality may matter the most, where for a one-step-ahead forecast short-term dynamics may be the dominating factor.
Thus, a one-step-ahead forecast may have totally different characteristics, in terms of possible accuracy, useful input features, well-performing methods, than a 400-step-ahead forecast~\citep{PETROPOULOS2014152}. 


Another requirement of OOS forecast evaluation is to make the forecast error measures insensitive to specific phases of business~\citep{TASHMAN2000437}. However, with a fixed origin setup, the errors may be the result of particular patterns only observable in that particular region of the horizon~\citep{TASHMAN2000437}, and evaluations will not generalise well to other phases such as Christmas sales, Holiday seasons, etc.. 
This poses a gap between what practitioners are really interested in and how forecasting is done often in academic settings. 
Having multiple forecasts for the same forecast step allows to produce a forecast distribution per each step for further analysis. Therefore, the following multi period evaluation setups are introduced as opposed to the fixed origin setup.


\subsection{Rolling Origin, Time Series Cross-Validation and Prequential Evaluation Setups}
\label{sec:rolling_origin_setup}

\citet{Armstrong1972} are among the first researchers to give a descriptive explanation of the rolling origin evaluation setup. 
Although the terms rolling origin setup and tsCV are used interchangeably in the literature, in addition to the forecast origin rolling forward, tsCV also allows to skip origins, effectively rolling forward by more than one step at a time (analogously to the difference between a leave-one-out CV and a k-fold CV). 
In the field of stream data mining and concept drift, this form of evaluation is furthermore known as \textit{interleaved-test-then-train} or \textit{prequential evaluation} \citep{gama2013evaluating, Ghomeshi2019} as means of online evaluation. For streaming data, prequential evaluation allows to continuously monitor the performance of a model that evolves over time, and thus detect and act upon concept drifts~\citep{10.1145/1557019.1557060, Kiran}.
With such multi period evaluations, each time the forecast origin updates, the model encounters new actual data. Hence, a rolling origin setup is typically the more practical evaluation setup in real-world application. For instance, in a task of forecasting the daily sales of a particular product, the number of actual sales can be obtained at the end of each day and can be incorporated in the model to better predict the next day's demand. 

With new data becoming available, we have the options to -- in the terminology of \cite{TASHMAN2000437} -- either update the model or recalibrate it. Recalibration here refers to either retraining the model weights from scratch or incremental learning as new data comes in. Updating on the other hand means just using the trained model to predict with new data. 
Although for some of the traditional models such as Exponential Smoothing (ETS) and Auto-Regressive Integrated Moving Average (ARIMA), the usual practice (and the implementation in the \texttt{forecast} package) in a rolling origin setup is to recalibrate (refit) the models, for general ML models it is more common to mostly just accept new data as inputs and only periodically retrain the model. While this is quite straightforward with a stateless ML model, on a model with a state such as a Recurrent Neural Network (RNN), updating still requires stepping through the whole series to construct the state. In this sense, the (updating-based) rolling origin setup comes more natural to many ML methods than to the traditional forecasting methods. Also, as ML methods tend to work better with higher granularities, re-fitting is not an option (for example, a monthly series predicted with ETS vs. a 5-minutely series predicted with Light Gradient Boosting Models). Therefore, retraining as the most recent data becomes available happens in ML methods mostly only when some sort of concept drift (change of the underlying data generating process) is encountered~\citep{webb_concept_drift}.


\begin{figure*}[htbp!]
	\centering
	\includegraphics[scale=1]{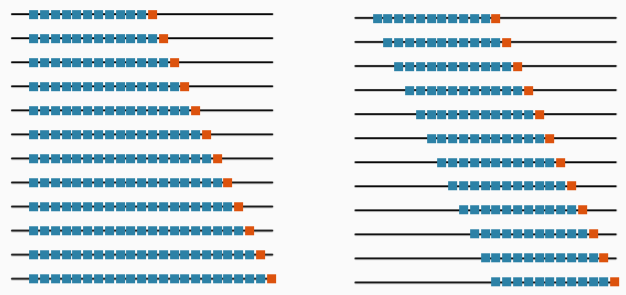}
	\caption{Comparison of Expanding Window vs. Rolling Window setups. The blue and orange points represent the training and test sets, respectively. The figure on the left side shows the Expanding Window setup where the training set keeps expanding. The figure on the right shows the Rolling Window setup where the size of the training set keeps constant and the first point of the training set keeps rolling forward.}
	\label{fig:rolling_methods}
\end{figure*}
Rolling origin evaluation can be conducted in two ways; 1) Expanding window setup 2) Rolling window setup. Figure \ref{fig:rolling_methods} illustrates the difference between the two approaches. In the expanding window setup, the training region of the series for the model expands as the forecast origin rolls forward, thus effectively increasing the length of the training data per each evaluation. The expanding window method is a good setup for small datasets/short series~\citep{bell_smyl_2019}. However, in the rolling window setup, the size of the training region is kept constant; thus as the forecast origin rolls forward, so does the start of the training period, dropping the oldest observations as new data becomes available~\citep{Cerqueira2020-bx}. The rolling window setup removes the oldest data from training. This will not make a difference with forecasting techniques that only minimally attend the distant past, such as ETS, but may be beneficial with pure autoregressive ML models, that have no notion of time beyond the windows. In a streaming data context as well, the most common methods of performing prequential evaluation are by using a sliding window or by using fading factors which tend to forget instances in the further past and focus on the current window~\citep{mulinka:hal-01952211}. This is because the usual prequential evaluation approach with expanding window is known to provide overestimation for the validation error~\citep{10.1145/1557019.1557060}. \citet{https://doi.org/10.1111/coin.12208} have empirically demonstrated that prequential evaluation with sliding window is the best approach for validation in comparison to fading factors and expanding window. A potential problem of the rolling origin setup is that the first folds may not have much data available. 
However, the size of the first folds is not an issue when dealing with long series, thus making rolling origin setup a good choice with sufficient amounts of data. On the other hand, with short series it is also possible to perform a combination of the aforementioned two rolling origin setups where we start with an expanding window setup and then move to a rolling window setup. 

\subsection{(Randomised) Cross-Validation}

\begin{figure*}[htbp!]
	\centering
	\includegraphics[width=0.6\textwidth]{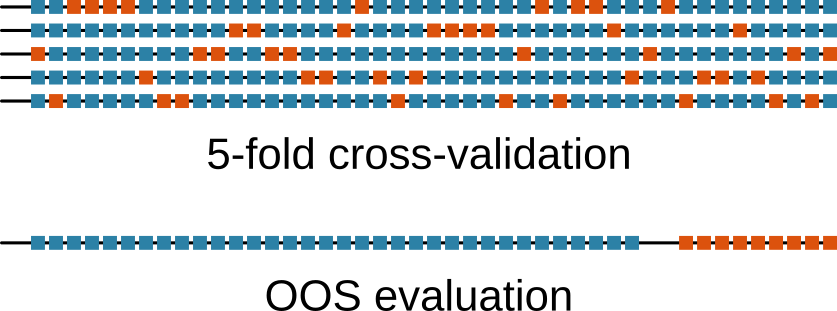}
	\caption{Comparison of randomised CV vs. OOS evaluation. The blue and orange dots represent the training and test sets, respectively. In the usual k-fold-CV setup the testing instances are chosen randomly over the series. In OSS, the test set is always reserved from the end of the series.}
	\label{fig:cross_validation}
\end{figure*}

The aforementioned two techniques of data partitioning preserve the temporal order of the time series when splitting and using the data.
Another form of data partitioning is to use a common randomised CV scheme as first proposed by \citet{stone1974}. This scheme is visualised in Figure \ref{fig:cross_validation}.
The dataset is initially shuffled randomly and then  partitioned into non-overlapping train and validation sets. Often a k-fold CV scheme is used for this purpose. For example, in a 5-fold CV strategy, the whole training dataset is randomly split into 5 partitions and 4 of them are used to train the model and the remaining partition held out for validation. This is done in iterations until all partitions are considered for validation separately. The set of validation scores produced this way are finally summarised. Leave-One-Out-Cross-Validation (LOOCV) is the extreme case of the k-fold CV where k is equal to the number of data points in the dataset. Therefore, compared to the aforementioned validation schemes which preserve the temporal order of the data, this form of randomised CV strategy can make efficient use of the data, since all the data is used for both model training as well as evaluation in iterations~\citep{Hastie2009-ft}. This helps to make a more informed estimation about the generalisation error of the model.


However, this form of random splitting of a time series does not preserve the temporal order of the data, and is therefore oftentimes not used and seen as problematic. The common points of criticism for this strategy are that, 1) it can make it difficult for a model to capture serial correlation between data points (autocorrelation) properly, 2) potential non-stationarities in time series can cause problems (for example, depending on the way that the data is partitioned, if all data from Sundays happen to be in the test set but not the training set in a series with weekly seasonality, then the model will not be able to produce accurate forecasts for Sundays since it has never seen data of Sundays before), 3) the training data contains future observations and the test set contains past data due to the random splitting and 4) since evaluation data is reserved randomly across the series, the forecasting problem shifts to a missing value imputation problem which certain time series models are not capable of handling~\citep{petropoulos2020forecasting}.

Out of these problems, to address the serial correlation issues, researchers have proposed a few variations of CV. Different forms of blocked CV have been introduced for this purpose, where the folds are selected in blocks, without initial shuffling and preserving the temporal order of the data within the folds~\citep{RACINE200039,BERGMEIR2012192}. Further non-dependent CV techniques have also been proposed, where a sufficiently sized block of instances surrounding the test instances are discarded to ensure independence between the training and test sets~\citep{non-dependent-cv, RACINE200039}. Nonetheless, randomised CV can be applied to pure AR models without a problem. \citet{BERGMEIR201870} theoretically and empirically show that CV performs well in a pure AR setup, as long as the models nest or approximate the true model, as then the errors are uncorrelated, leaving no dependency between the individual windows. To check this, it is important to estimate the serial correlation of residuals. For this, the Ljung-Box test~\citep{Ljung-box} can be used on the OOS residuals of the models. While for overfitting models there will be no autocorrelation left in the residuals, if the models are underfitted, some autocorrelation will be left in the OOS residuals. If there is autocorrelation left, then the model still does not use all the information available in the data, which means there will be dependencies between the separate windows. In such a scenario, CV of the time series dataset will not hold valid, and underestimate the true generalisation error.
The existence of significant autocorrelations anyway means that the model should be improved to do better on the respective series (increase the AR order to capture autocorrelation etc.), since the model has not captured all the available information. Once the models are sufficiently competent in capturing the patterns of the series, for pure AR setups (without exogenous variables), standard k-fold CV is a valid strategy. Therefore, in situations with short series and small amounts of training data, where it is not practically feasible to apply the aforementioned tsCV techniques due to the initial folds involving very small lengths of the series, the standard CV method with some control of underfitting of the models is a better choice with efficient use of data. 

The aforementioned problem that the testing windows can contain future observations, is also addressed by \citet{BERGMEIR201870}. With the CV strategy, the past observations not in the training data but existing in the test set can be considered missing observations, and the task is seen more as a missing value imputation problem rather than a forecasting problem. Many forecasting models such as ETS (in its implementation in the \texttt{forecast} package~\citep{robgeorg2018otext}), which iterate throughout the whole series, cannot properly deal with missing data. For RNNs as well, due to their internal states that are propagated forward along the series, standard k-fold CV which partitions data randomly across the series is usually not applicable. Therefore, for such models, the only feasible validation strategy is tsCV. Models such as ETS can anyway train competitively with minimal amounts of data (as is the case with the initial folds of the tsCV technique) and thus, are not quite problematic with tsCV. However, for reasonably trained pure AR models, where the forecasts for one window do not in any way depend on the information from other windows (due to not underfitting and having no internal state), it does not make a difference between filling the missing values in the middle of the series and predicting future values, where both are performed OOS. Nevertheless, the findings by \citet{BERGMEIR201870} are restricted to only stationary series. \citet{Cerqueira2020-bx}'s work on the same area has concluded that for stationary series, using a pure AR setup, a blocked CV strategy works the best, and they also perform an analysis on non-stationary data, discussed in the following section.

\subsection{Data partitioning for non-stationary data}

\citet{Cerqueira2020-bx} experimented using non-stationary series, where they have concluded that OOS validation procedures preserving the temporal order (such as tsCV), are the right choice when non-stationarities exist in the series. However, a possible criticism of that work is the choice of models. We have seen in Section~\ref{sec:motivation} that ML models are oftentimes not able to address certain types of non-stationarities out of the box. More generally speaking, ML models are non-parametric, data-driven models. As such, the models are typically very flexible and the function fitted depends heavily on the characteristics of the observed data. Though recently challenged \citep{balestriero2021learning}, a common notion is that ML models are typically good at interpolation and lack extrapolation capabilities.
The models used by \citet{Cerqueira2020-bx} include several ML models such as a Rule-based Regression (RBR) model, a Random Forest (RF) model and a Generalised Linear Model (GLM), without in any way explicitly tackling the non-stationarity in the data (similar to our example in Section~\ref{sec:motivation}). Thus, if a model is poor and not producing good forecasts, performing a validation to select hyperparameters, using any of the aforementioned CV strategies, will be of limited value. Furthermore, and more importantly, non-stationarity is a broad concept and it will depend both for the modelling and the evaluation on the type of non-stationarity which procedures will perform well. For example, with abrupt structural breaks and level shifts occurring in the unknown future, but not in the training and test set, it will be impossible for the models to address this change and none of the aforementioned evaluation strategies would do so either. In this situation, even tsCV would grossly underestimate the generalisation error. For a more gradual underlying change of the DGP, a validation set at the end of the series would be more appropriate since in that case, the data points closer to the end of the series may be already undergoing the change of the distribution. On the other hand, if the series has deterministic trend or seasonality, which are straightforward to forecast, they can be simply extracted from the series and predicted separately whereas the stationary remainder can be handled using the model. In such a setup, the k-fold CV scheme will work well for the model, since the remainder complies with the stationarity condition. For other non-deterministic trends, there are several data pre-processing steps mentioned in the literature such as lag-1 differencing, logarithmic transformation (for exponential trends), Seasonal and Trend Decomposition using Loess (STL Decomposition), local window normalisation~\citep{HEWAMALAGE2021388}, moving average smoothing, percentage change transform, wavelet transform etc.~\citep{Salles2019-ql}. \citet{Salles2019-ql} have conducted an extensive empirical study to investigate the impact of the choice of the data transformation technique on the accuracy of the model, using a linear ARMA model. Their findings have concluded that there is no single universally best transformation technique across all datasets; rather it depends on the characteristics of the individual datasets. However, for the particular datasets used in their study, differencing and moving average smoothing have generally worked the best for addressing trend. If appropriate data pre-processing steps are applied to enable models to handle non-stationarities, with a pure AR setup, the CV strategy still holds valid after the data transformation, if the transformation achieves stationarity. As such, to conclude, for non-stationarities, tsCV seems the most adequate as it preserves the temporal order in the data. However, there are situations where also tsCV will be misleading, and the forecasting practitioner will already for the modeling need to attempt to understand the type of non-stationarity they are dealing with. This information can subsequently be used for evaluation, which may render CV methods for stationary data applicable after transformations of the data to make them stationary.

\subsection{Other model selection methods}

The aforementioned data partitioning schemes are quite important when it comes to model selection based on the performance of the models on the validation sets.
Apart from the CV strategies discussed above, other model selection techniques exist such as information criteria (IC) or techniques like minimum message length \citep{fitzgibbon2004minimum}. The advantage of these techniques over data partitioning is typically that all data can be used for training and no validation set is needed. 

For example, let us consider Akaike's Information Criterion \citep[AIC, ][]{akaike1974new}, but similar considerations hold for other IC.  
For time series models, it has been found that minimising AIC is asymptotically equivalent to minimising the MSE of OOS one-step ahead forecasts~\citep{INOUE2006273}.
However, using AIC has several downsides. To use it to compare across models, the respective likelihoods need to be computed the same way, using the same data. Therefore, it cannot be used to compare across models, with different model orders, from different model families such as ETS and ARIMA (since likelihoods for those models are computed in different ways), with and without differencing, since differencing effectively reduces the amount of data points available for the model. Apart from that, AIC is in general used when getting access to a separate test set is expensive (due to limited data), which is often not the case with data-abundant scenarios where ML models are applicable. Therefore, AIC is more suitable for small datasets and this is why models such as ETS and ARIMA that are not very data-intensive, use AIC and other IC internally. Moreover, the definition of AIC uses (an estimation of) the number of parameters of the model, which is not straightforward for complex ML models, since simply counting the number of parameters does not represent well the complexity of such models. Due to these reasons, generally AIC and other IC are not used for ML models that work on large datasets. For such situations, the data partitioning methods discussed before are usually preferable.





\subsection{Summary and guidelines for data partitioning and model selection}

It is important to identify which out of the above data partitioning strategies most closely estimates (without under/overestimation) the final error of a model for the test set under the given scenario (subject to different non-stationarities/serial correlations/amount of data of the given time series). In particular, in the M5 competition as well, it was re-emphasised that a reliable CV strategy is essential, to be able to assess the generalisation error of models~\citep{makridakism5}.

\begin{figure*}[htbp!]
	\hspace{-0.5cm}
	\includegraphics[scale=0.28]{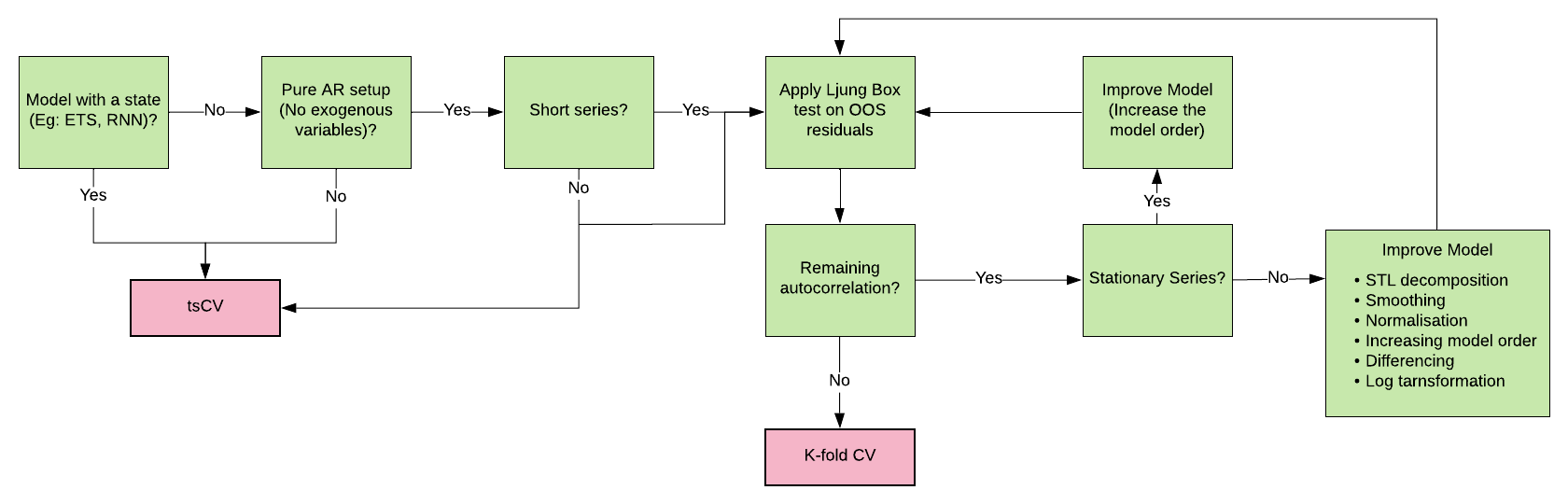}
	\caption{Guidelines on Data Partitioning and Model Selection}
	\label{ch7:fig:model_selection_flowchart}
\end{figure*}

The gist of the guidelines for model selection is visualised by the flow chart in Figure \ref{ch7:fig:model_selection_flowchart}. If the series are not short, tsCV is usually preferrable over k-fold CV, if there are no practical considerations such as that an implementation of an algorithm is used that is not primarily intended for time series forecasting, and that internally performs a certain type of cross-validation. If series are short, then k-fold CV should be used, accounting adequately for non-stationarities and autocorrelation in the residuals.

\section{Error Measures for Forecast Evaluation}
\label{sec:error_measures}



Once the predictions are obtained from models, the next requirement is to compute errors of the predictions to assess the model performance. \citet{Belt2017} argue in their work that a bias-variance decomposition of the error measure should be considered. Bias and variance of forecasts may yield different business decisions.
Bias in predictions occurs due to errors from wrong model assumptions, which result in a weak model not having captured the exact patterns of the data. This happens mostly due to the selected sample of data (used for model training) being under-representative of the whole distribution, which is also called as the selection bias. Because of this reason, a model can be very accurate (forecasts being very close to actuals), but consistently produce more overestimations than underestimations, which may be concerning from a business perspective. Therefore, forecast bias is calculated with a sign, as opposed to absolute errors, so that it indicates the direction of the forecast errors, either positive or negative. For example, scale-dependent forecast bias can be assessed with the Mean Error (ME) as defined in Equation in \ref{eqn:bias}. Here, $y_t$ indicates the true value of the series, $\hat{y}_t$ the forecast and $n$, the number of all available errors (across series, across horizons, etc.).
The scale-dependent standard deviation (Std) of the errors for the population is defined in Equation \ref{eqn:std}, assuming a 0 population mean of the errors, i.e., an unbiased model. 
Note that the Std of the errors for an unbiased model is identical to the Root Mean Squared Error (RMSE) defined later in Equation \ref{eqn:rmse}. Therefore, RMSE produces an estimate of the Std of the distribution of forecast errors.
Other scale-free versions of bias and Std can be defined by scaling with respect to appropriate scaling factors, such as actual values of the series.

\begin{equation}
	\label{eqn:bias}
	\textit{ME} = \frac{1}{n}\sum_{t=1}^{n} (y_t - \hat{y_t})
\end{equation}

\begin{equation}
	\label{eqn:std}
	\textit{Error Std} = \sqrt{\frac{1}{n}\sum_{t=1}^{n} (y_t - \hat{y_t})^2}
\end{equation}

Two other popular and simple error measures used in a usual regression context are Mean Squared Error (MSE) and Mean Absolute Error (MAE) defined in Equations \ref{eqn:mse} and \ref{eqn:mae} respectively.

\begin{equation}
	\label{eqn:mse}
	\textit{MSE} = \frac{1}{n}\sum_{t=1}^{n} (y_t - \hat{y_t})^2
\end{equation}

\begin{equation}
	\label{eqn:mae}
	\textit{MAE} = \frac{1}{n}\sum_{t=1}^{n} |y_t - \hat{y_t}|
\end{equation}





Apart from these simple measures used as in usual ML tasks such as regression, for forecasting a wide variety of error measures have been proposed by researchers over the years. The main reason for this is the need to have measures that allow for comparisons across series. For this, the measures need to be scaled, and it has turned out to be next to impossible to develop a scaling procedure that works for any type of possible non-stationarity and non-normality in a time series. 
Eventually we encounter a particular condition of the time series in the real world, that makes the proposed error measure fail~\citep{SvetunkovAdam}. 
Thus, new measures usually either focus on specific business needs or have the intention of addressing issues of previously used measures, but have new issues then.
This has led to a large pool of proposals in the literature. Researchers/practitioners very often have particular series in mind when evaluating in a certain way (using specific measures) in their work, but these preconditions are often never stated. For example, smart meter or wind power production series do not usually have exponential trends, and they hardly have level shifts or long-term trends etc. On the other hand, growing businesses such as tech start-ups or ride-share providers often have strong trends in any business related time series that they have collected. 
The key to selecting a particular error measure for forecast evaluation is that it is mathematically and practically robust under the given data. From a business point of view there can be other requirements for an error measure such as being interpretable (easy to communicate) and reflecting on the key performance indicators of the underlying business application such as the net profit; which we do not focus on in this work.



Point forecasting which is the main focus of this work is about predicting a particular statistic of interest from the future distribution of values, such as mean or median. Therefore, different point forecast evaluation measures are also targeted towards optimising for a specific statistic of the distribution and it is important to distinguish which statistic it is for each error measure. For example, measures with squared base errors such as MSE and RMSE optimise for the mean whereas others with absolute value base errors such as MAE and Mean Absolute Scaled Error (MASE) optimise for the median. 
Although the mean and median are the same for a symmetric distribution, that does not hold for skewed distributions as with intermittent series. There exist numerous controversies in the literature regarding this. \citet{petropoulos2020forecasting} suggest that it is not appropriate to evaluate the same forecasts using many different error measures, since each one optimises for a different statistics of the distribution. Also according to \citet{KOLASSA2020208}, if different point forecast evaluation measures are considered, multiple point forecasts for each series and time point also need to be created. \citet{KOLASSA2020208} further argues that, if the ultimate evaluation measure is, e.g., MAE which focusses on the median of the distribution, it does not make sense to optimise the models using an error measure like MSE (which accounts for the mean). It is more meaningful to consider MAE also during model training as well. However, these arguments hold only if it is not an application requirement for the same forecasts to perform generally well under all these measures. \citet{Koutsandreas2021-ww} have empirically shown that, when the sample size is large, a wide variety of error measures agree on the most consistently dominating methods as the best methods for that scenario. They have also demonstrated that using two different error measures for optimising and final evaluation has an insignificant impact on the final accuracy of the models. \citet{Bermudez2006-md} have developed a fuzzy ETS model optimised via a multi-objective function combining three error measures MAPE, RMSE and MAE. Empirical results have demonstrated that using such a mix of error measures instead of just one for the loss function leads to overall better, robust and generalisable results even
when the final evaluation is performed with just one of those measures. \citet{google_session_2020} also assess their same forecasts across numerous error measures in a business context.

On the other hand, \citet{KOLASSA2016788} also argues that point forecast evaluation alone is not sufficient. This is due to all the pitfalls associated with every point forecast evaluation measure as discussed in the rest of this section. Also, according to that author, evaluating only individual statistics of a distribution is not adequate and the overall predictive distributions need to be considered instead, in order to estimate the uncertainty of the produced point forecasts with a reasonable confidence. Although not the main point of focus in this article, different evaluation measures have been proposed in this respect. \citet{KOLASSA2016788} explains most of them including randomised probability integral transform (rPIT) and several proper scoring rules such as logarithmic score, Brier score, ranked probability score etc.

Furthermore, depending on different data partitioning schemes, we may obtain many errors for the same model either for a fixed forecast origin or rolling forecast origin etc. In the context of global forecasting models, we train using many different series and also often evaluate on that same set of series. Thus, the number of series further increases the amount of errors available for a single model. For summarising errors across all available series as well as the different steps in the forecast horizon, we can consider a number of statistics such as median, arithmetic mean or geometric mean etc. Using medians and geometric means instead of arithmetic means helps with avoiding sensitivity to outlier series in a set of time series. However, this means that using the arithmetic mean identifies the existence of such outlier errors for certain series. The problem specifically with geometric mean based measures is that a model can perform perfectly (0 or $<1$ error) on one series and quite bad on all the others but still become the best (as the overall error becomes 0 or quite small due to multiplication)~\citep{Boylan2006}. Thus, \citet{SvetunkovAdam} mentions that the use of several summary operators on the same errors can raise awareness regarding these issues of the individual errors. Summarising across series and across the horizon can be done using the same or different statistical operators. We can also change the order of summarising the errors; for instance we can first summarise the errors for the different time steps separately and then summarise those per time step error measures. When summarising across the horizon, weights can be assigned to different time steps to get a weighted measure as well. Different terminology has been introduced by researchers for these base errors, statistical operators etc. For instance, in the work by \citet{kunst}, the base error functions are named as local distance functions and the statistical operators to summarise base errors are denoted as link functions. The resulting final error functions are known as metrics or distance functions. According to the terminology introduced by \citet{hyndman2006evaluation}, which we follow in this paper, the base errors are called Errors, and summarised by using different statistical operators into Error Measures. 

There are many different point forecast error measures available in the forecasting literature categorised based on 1) whether squared or absolute errors are used 2) techniques used to make them scale-free and 3) the operator such as mean, median used to summarise the errors~\citep{Koutsandreas2021-ww}.  
%
%
In the rest of this section, we first introduce, define, and categorise all different error measures introduced in the literature. We also provide the formulae for these measures and their general mathematical issues (not relevant to specific characteristics of the series). Then, we move on to provide an overview of certain types of data and possible problems that the error measures can have with them. We discuss which error measures are preferrable or should be avoided depending on each of the characteristics of time series as also stated in Section \ref{sec:characteristics_of_timeseries}.
	
\begin{figure*}[htbp!]
	\vspace*{-1.6cm}
	\centering
	\includegraphics[scale=0.24]{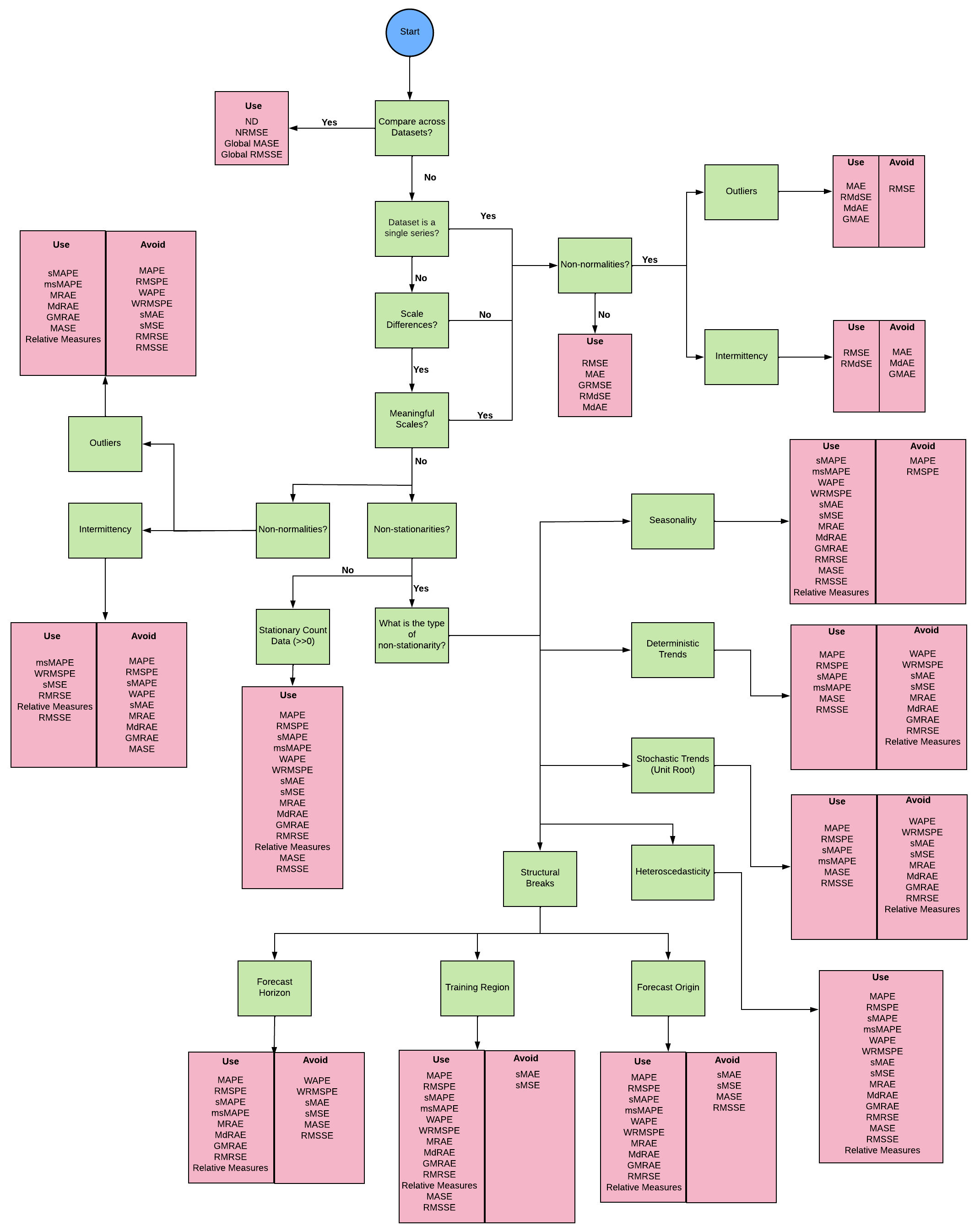}
	\caption{Flow Chart for Forecast Error Measure Selection}
	\label{fig:error_measure_selection_flow_chart}
\end{figure*}

Summarising all these details, the main results of this section are then Table~\ref{tab:checklist_error_selection} and Figure~\ref{fig:error_measure_selection_flow_chart}. Table \ref{tab:checklist_error_selection} can be used to choose error measures under given characteristics of the data.
In Table \ref{tab:checklist_error_selection}, the scaling column indicates the type of scaling associated with each error measure mentioned in the previous column. This includes no scaling, scaling based on actual values, scaling based on benchmark errors as well as the categorisation such as per-step, per-series and all-series (per-dataset) scaling. The $\dagger$ sign in Table \ref{tab:checklist_error_selection} indicates that the respective error measures need to be used with caution under the given circumstances as explained in the above discussions. The flow chart in Figure \ref{fig:error_measure_selection_flow_chart} provides further support for forecast evaluation measure selection based on user requirements and other characteristics in the data. In Figure \ref{fig:error_measure_selection_flow_chart}, the error measures selected to be used with outlier time series are in the context of being robust against outliers, not capturing them.

\begin{sidewaystable}[]
	\caption{Checklist for selecting error measures for final forecast evaluation based on different time series characteristics}
	\scalebox{0.75}{
		\begin{tabular}{lllcccccccccc}
			\hline
			\multirow{2}{*}{\begin{tabular}[c]{@{}c@{}}Stationary Count\\ Data (\textgreater{}\textgreater{}0)\end{tabular}} & \multirow{2}{*}{Seasonality} & \multirow{2}{*}{\begin{tabular}[c]{@{}c@{}}Trend\\ (Linear/Exp.)\end{tabular}} & \multirow{2}{*}{\begin{tabular}[c]{@{}c@{}}Unit\\ Roots\end{tabular}} & \multirow{2}{*}{Heteroscedasticity} & \multicolumn{3}{c}{\begin{tabular}[c]{@{}c@{}}Structural Breaks\\ (With Scale Differences)\end{tabular}}                                                                          & \multirow{2}{*}{Intermittence} & \multirow{2}{*}{Outliers} & \multirow{2}{*}{Error Measures} & \multicolumn{2}{c}{\multirow{2}{*}{Scaling}} \\ \cline{6-8}
			\multicolumn{2}{c}{}                                                                                                                                                                                                                                                                             &                                                                                &                                                                       &                                     & \begin{tabular}[c]{@{}c@{}}Forecast\\ Horizon\end{tabular} & \begin{tabular}[c]{@{}c@{}}Training\\ Region\end{tabular} & \begin{tabular}[c]{@{}c@{}}Forecast\\ Origin\end{tabular} &                                &                        & &  & \\ \hline
			\cmark                                                                                               & \cmark                      & \cmark                                                                        & \cmark                                                               & \cmark                             & \cmark                                                    & \cmark                                                   & \cmark                                                   & \cmark                        & \xmark     &         RMSE &     \multicolumn{2}{c}{\multirow{2}{*}{None}}    \\ \cline{1-11} 
			\cmark                                                                                               & \cmark                      & \cmark                                                                        & \cmark                                                               & \cmark                             & \cmark                                                    & \cmark                                                   & \cmark                                                   & \xmark                          & \cmark         & MAE &     \multicolumn{2}{c}{}     \\ \hline
			\cmark                                                                                               & \xmark                        & \cmark                                                                        & \cmark \DAG                                                       & \cmark \DAG                     & \cmark                                                    & \cmark                                                   & \cmark                                                   & \xmark                          & \xmark         & MAPE &   \multirow{4}{*}{\begin{tabular}[c]{@{}l@{}}OOS\\ Per Step\end{tabular}} & \multirow{10}{*}{\begin{tabular}[c]{@{}l@{}}Actual\\ Values\end{tabular}}          \\ \cline{1-11} 
			\cmark                                                                                               & \xmark                        & \cmark                                                                        & \cmark \DAG                                                       & \cmark \DAG                     & \cmark                                                    & \cmark                                                   & \cmark                                                   & \xmark                          & \xmark    & RMSPE & &                 \\ \cline{1-11} 
			\cmark                                                                                               & \cmark                      & \cmark                                                                        & \cmark                                                               & \cmark                             & \cmark                                                    & \cmark                                                   & \cmark                                                   & \xmark                          & \cmark     &  sMAPE                           & &      \\ \cline{1-11} 
			\cmark                                                                                               & \cmark                      & \cmark                                                                        & \cmark                                                               & \cmark                             & \cmark                                                    & \cmark                                                   & \cmark                                                   & \cmark                        & \cmark     & msMAPE                          &  &              \\ \cline{1-12} 
			\cmark                                                                                               & \cmark                      & \xmark                                                                          & \xmark                                                                 & \cmark                             & \xmark                                                      & \cmark                                                   & \cmark                                                   & \xmark                          & \xmark      & WAPE         & \multirow{2}{*}{\begin{tabular}[c]{@{}l@{}}OOS\\ Per Series\end{tabular}} &      \\ \cline{1-11} 
			\cmark                                                                                               & \cmark                      & \xmark                                                                          & \xmark                                                                 & \cmark                             & \xmark                                                      & \cmark                                                   & \cmark                                                   & \cmark \DAG                & \xmark     & WRMSPE &&                \\ \cline{1-12} 
			\cmark                                                                                               & \cmark                      & \xmark                                                                          & \xmark                                                                 & \cmark                             & \xmark                                                      & \xmark                                                     & \xmark                                                     & \xmark                          & \xmark        & sMAE                            &  \multirow{2}{*}{\begin{tabular}[c]{@{}l@{}}In-Sample\\ Per Series\end{tabular}} &             \\ \cline{1-11} 
			\cmark                                                                                               & \cmark                      & \xmark                                                                          & \xmark                                                                 & \cmark                             & \xmark                                                      & \xmark                                                     & \xmark                                                     & \cmark \DAG                & \xmark   & sMSE                            & &                  \\ \cline{1-12} 
			\cmark                                                                                               & \cmark                      & \cmark                                                                        & \cmark                                                               & \cmark                             & \cmark                                                    & \cmark                                                   & \cmark                                                   & \xmark                          & \cmark          & ND                              &  \multirow{2}{*}{\begin{tabular}[c]{@{}l@{}}OOS\\ All Series\end{tabular}}    &       \\ \cline{1-11} 
			\cmark                                                                                               & \cmark                      & \cmark                                                                        & \cmark                                                               & \cmark                             & \cmark                                                    & \cmark                                                   & \cmark                                                   & \cmark                        & \xmark        & NRMSE                           & &           \\ \hline
			\cmark \DAG                                                                                       & \cmark \DAG              & \xmark                                                                          & \xmark                                                                 & \cmark                             & \cmark                                                    & \cmark \DAG                                           & \cmark                                                   & \xmark                          & \cmark \DAG    & MRAE & \multirow{4}{*}{\begin{tabular}[c]{@{}l@{}}OOS\\ Per Step\end{tabular}} & \multirow{8}{*}{\begin{tabular}[c]{@{}l@{}}Benchmark\\ Errors\end{tabular}}       \\ \cline{1-11} 
			\cmark \DAG                                                                                       & \cmark \DAG              & \xmark                                                                          & \xmark                                                                 & \cmark                             & \cmark                                                    & \cmark \DAG                                           & \cmark                                                   & \xmark                          & \cmark \DAG  & MdRAE &&         \\ \cline{1-11} 
			\cmark \DAG                                                                                       & \cmark \DAG              & \xmark                                                                          & \xmark                                                                 & \cmark                             & \cmark                                                    & \cmark \DAG                                           & \cmark                                                   & \xmark                          & \cmark \DAG  & GMRAE &&         \\ \cline{1-11} 
			\cmark \DAG                                                                                       & \cmark \DAG              & \xmark                                                                          & \xmark                                                                 & \cmark                             & \cmark                                                    & \cmark \DAG                                           & \cmark                                                   & \cmark \DAG                & \xmark  & RMRSE &&                    \\ \cline{1-12} 
			\cmark \DAG                                                                                       & \cmark \DAG              & \xmark                                                                          & \xmark                                                                 & \cmark                             & \cmark                                                    & \cmark \DAG                                           & \cmark                                                   & \cmark \DAG                & \cmark \DAG    & Relative Measures & \begin{tabular}[c]{@{}l@{}}OOS\\ Per Series\end{tabular} &       \\ \cline{1-12} 
			\cmark \DAG                                                                                       & \cmark \DAG              & \cmark                                                                        & \cmark                                                               & \cmark                             & \xmark                                                      & \cmark \DAG                                           & \xmark                                                     & \xmark                          & \cmark     & MASE & \multirow{2}{*}{\begin{tabular}[c]{@{}l@{}}In-Sample\\ Per Series\end{tabular}} &              \\ \cline{1-11} 
			\cmark \DAG                                                                                       & \cmark \DAG              & \cmark                                                                        & \cmark                                                               & \cmark                             & \xmark                                                      & \cmark \DAG                                           & \xmark                                                     & \cmark \DAG                & \xmark       & RMSSE                           & &              \\ \cline{1-12} 
			\cmark \DAG                                                                                       & \cmark \DAG              & \cmark                                                                        & \cmark                                                               & \cmark                             & \cmark                                                    & \cmark                                                   & \cmark                                                   & \cmark                        & \cmark   & & \begin{tabular}[c]{@{}l@{}}In-Sample\\ All Series\end{tabular}  &                \\ \hline
			\cmark                                                                                               & \cmark                      & \cmark \DAG                                                                & \cmark                                                               & \cmark \DAG                     & \cmark                                                    & \cmark                                                   & \cmark                                                   & \cmark                        & \cmark        & \begin{tabular}[c]{@{}l@{}}Measures with\\ Transformations\end{tabular} & \multicolumn{2}{c}{None}           \\ \hline
	\end{tabular}}
	\label{tab:checklist_error_selection}
\end{sidewaystable}

\subsection{Categorisation of Error Measures}
\label{sec:error_measures_categorisation}
Different error measures in the literature are categorised as detailed in the following. We follow the categorisation of error measures introduced by \citet{hyndman2006evaluation} and adopted by many successive works in this space.

\subsubsection{Scale-Dependent Error Measures} 

Scale-dependent measures as the name suggests, are dependent on the scale of the series. In all the scale-dependent measures defined below, the scale-dependent base error $e_t$ used is as defined in Equation \ref{eqn:scale_dependent_error}. 

\begin{equation}
	\label{eqn:scale_dependent_error}
	e_t = y_t - \hat{y}_t
\end{equation}

Apart from MSE and MAE as defined in Equations \ref{eqn:mse} and \ref{eqn:mae}, other examples of scale-dependent measures which use $e_t$ as the base error are as defined below. 


\begin{enumerate}
	\item Root Mean Squared Error (RMSE)
	\begin{equation}
		\label{eqn:rmse}
		\textit{RMSE} = \sqrt{\frac{1}{n}\sum_{t=1}^{n}(e_t^2)}
	\end{equation}
	\item Root Median Squared Error (RMdSE)
	\begin{equation}
		\textit{RMdSE} = \sqrt{ \textit{median}(e_t^2)}
	\end{equation}
	\item Median Absolute Error (MdAE)
	\begin{equation}
		\textit{MdAE} =  \textit{median}(|e_t|)
	\end{equation}
	\item Geometric Root Mean Squared Error (GRMSE) - Proposed by \citet{SYNTETOS2005303}. 
	\begin{equation}	
		\textit{GRMSE} = \sqrt[2n]{\prod_{t=1}^{n}{e_t^2}}
	\end{equation}
	
	Geometric Mean Absolute Error (GMAE)
	\begin{equation}
		\textit{GMAE} = \sqrt[n]{\prod_{t=1}^{n}{|e_t|}}
	\end{equation}
\end{enumerate}

Since RMSE is on the same scale as the original data (due to the square and the square-root), it is often preferred over MSE. Squared errors are known to lead to unbiased point forecasts since they predict the mean when used as a loss function~\citep{KOLASSA2016788}. Although the definitions are slightly different, mathematically both GMAE and GRMSE are equivalent. The problem specifically with these geometric mean based measures is that a model can perform perfectly (0 or $<1$ error) on one series and quite bad on all the others but still become the best (as the overall error becomes 0 or quite small due to multiplication)~\citep{Boylan2006}.

Scale-invariant measures on the other hand are introduced for the requirement to be able to compare across series having different scales. Traditionally in forecasting, one series was mostly considered as a single dataset by practitioners in the domain. Therefore, scaling errors was limited to scales computed per each series or even each time step (to be able to compare errors across series). These approaches had several issues as explained next. However, in the current context of Big Data and global forecasting models, we are now in the situation where we want to compare models across datasets or select models that perform generally well on many datasets each having many series. Consequently, the error measures being introduced have also shifted from computing a scale per-series to a scale for the whole dataset. Such global scaling based error measures can address some of the issues with the per-step or per-series scaling. Nevertheless, \citet{chen_twycross_garibaldi_2017} state that despite the type of scaling used, the resulting error measure values need to be closely related to the scale of the series at the specific observation points. Hence, those authors opt for those measures that compute a per-step scaling. Due to the non-stationarities and non-normalities that are inherent in many time series, a constant estimator for scale (along the series or for the whole dataset) has been shown as a comparatively poor form of scaling for time series~\citep{chen_twycross_garibaldi_2017}.

\subsubsection{Measures based on Percentage Errors}
\label{sec:percentage_measures}

In percentage error based measures, base errors are scaled by actual time series values. This means that the time series values get scaled with respect to the actual scale of the series. Percentage based measures were invented for inventory based series (having very high volumes, no intermittency) since they are more meaningful as indicating the percentage loss and thus easy to communicate. The percentage error is usually defined as in Equation \ref{eqn:percentage_error}, where $e_t$ is the scale-dependent error and $y_t$ is the actual value at the $t^{th}$ time step.

\begin{equation}
	\label{eqn:percentage_error}
	p_t = \frac{100e_t}{y_t}
\end{equation}
Examples of percentage errors are as follows.

\begin{enumerate}
	\item Mean Absolute Percentage Error (MAPE)
	\begin{equation}
		\textit{MAPE} = \frac{1}{n}\sum_{t=1}^{n}(|p_t|)
	\end{equation}
	\item Median Absolute Percentage Error (MdAPE)
	\begin{equation}
		\textit{MdAPE} =  \textit{median}(|p_t|)
	\end{equation}
	\item Root Mean Square Percentage Error (RMSPE) - Used in the Rossmann Store Sales Forecasting Competition\footnote{\url{https://www.kaggle.com/c/rossmann-store-sales}}~\citep{BOJER2020}
	\begin{equation}
		\textit{RMSPE} = \sqrt{\frac{1}{n}\sum_{t=1}^{n}(p_t^2)}
	\end{equation}
	\item Root Median Square Percentage Error (RMdSPE)
	\begin{equation}
		\textit{RMdSPE} = \sqrt{ \textit{median}(p_t^2)}
	\end{equation}
\end{enumerate}

Percentage based measures have the issue that $p_t$ is not symmetric since exchanging $y_t$ with $\hat{y}_t$ changes the value of the error measure. Moreover, percentages can be above 100\% sometimes, which makes a flat forecast of all 0's a better forecast with a 100\% MAPE~\citep{kolassa_2017}. Apart from that, percentage based measures only make sense with the existence of a meaningful zero (where divisions and ratios are meaningful). Other forms of percentage based measures have been introduced with different scaling factors.

\begin{enumerate}
	\setcounter{enumi}{4}
	\item Symmetric Mean Absolute Percentage Error (sMAPE) - Idea first proposed by \citet{MAKRIDAKIS1993527}\footnote{The original definition of sMAPE used at the M3 forecasting competition did not use absolute values of $y_t$ and $\hat{y}_t$ in the denominator, since all series had only positive values there.}
	\begin{equation}
		\textit{sMAPE} = \frac{1}{n}\sum_{t=1}^{n}(\frac{200|e_t|}{|y_t| + |\hat{y}_t|})
		\label{eqn:smape}
	\end{equation}
	\item Symmetric Median Absolute Percentage Error (sMdAPE)
	\begin{equation}
		\textit{sMdAPE} =  \textit{median}(\frac{200|e_t|}{|y_t| + |\hat{y}_t|})
	\end{equation}
\end{enumerate}



sMAPE fixes the type of asymmetry seen with MAPE that the penalisation is different if $y_t$ and $\hat{y}_t$ are exchanged. Hence, it was used heavily in most early forecasting competitions. However, sMAPE is not as symmetric as its name suggests \citep{GOODWIN1999405}; it is arguably even less symmetric. It penalises the underestimates more than the overestimates for the same value of $y_t$. Thus, it may tend towards selecting a slightly overestimating model. Nevertheless, this kind of asymmetry may be of interest to certain domains where underestimates are considered to be more costly than overestimates~\citep{Scott_Armstrong_undated}. sMAPE is also in general further criticised for its lack of interpretability~\citep{hyndman2006evaluation}.





\begin{enumerate}
	\setcounter{enumi}{6}
	\item Modified Symmetric Mean Absolute Percentage Error (msMAPE)
	\begin{equation}
		\textit{msMAPE} = \frac{1}{n}\sum_{t=1}^{n}\frac{200|e_t|}{ \textit{max}(|y_t| + |\hat{y}_t| + \epsilon, 0.5 + \epsilon)} 
		\label{eqn:msmape}
	\end{equation}
\end{enumerate}

This was introduced by \citet{Suilin_undated-hh}, to address issues with sMAPE, as detailed in the next Section. In Equation \ref{eqn:msmape}, $\epsilon = 0.1$ by default. msMAPE assigns the same scaling factor (0.6 in this case) for all $|y_t| + |\hat{y}_t|$ less than or equal a certain threshold (0.5 in this case). This means that in this range, all errors are simply divided by a fixed constant, irrespective of the actual value or the forecast of the series. This idea is similar to the concept of winsorising of error distributions suggested by \citet{ARMSTRONG199269}, which is to clip extreme values/outliers (tails of the distribution) of the errors and replace them with certain limits/thresholds. \citet{Arnottjfds.2019.1.064} discuss the same idea of winsorisation for the purpose of excluding outliers from the data in the first place. In this sense, msMAPE can be considered as a winsorised version of sMAPE.

However, the exact threshold at which to cut off the errors/data, depends on the scale of the series. For example, this threshold cannot be the same (say 0.5) on two series where one goes to a maximum value of 1000 and the other has values only in-between 0-1. Also, winsorising basically skews the error measure/data since all values in the distribution are cut off at a certain threshold. Therefore, msMAPE is rather ad-hoc and it does not estimate any statistic like the mean or median of the distribution; it is rather a biased version of mean/median. Apart from that, the issues arising from symmetry in sMAPE are still there in msMAPE as well, but only for larger actuals and forecasts that exceed the threshold. Due to these reasons, msMAPE is often disapproved by researchers, as it has no theoretical foundation and its statistical properties have not been explored.


\citet{KIM2016669} introduced the Mean Arctangent Absolute Percentage Error (MAAPE), which retains the original scaling factor of percentage based measures, and thus its associated intuitive interpretation.

\begin{enumerate}
	\setcounter{enumi}{7}
	\item Mean Arctangent Absolute Percentage Error (MAAPE)
	\begin{equation}
		\textit{MAAPE} = \frac{1}{n}\sum_{t=1}^{n} \textit{arctan}(|\frac{e_t}{y_t}|)
	\end{equation}
\end{enumerate}

The Weighted Absolute Percentage Error (WAPE) is defined as in Equation \ref{eqn:wape} and performs the scaling based on the OOS values of the series in the whole forecast horizon.

\begin{enumerate}
	\setcounter{enumi}{8}
	\item Weighted Absolute Percentage Error (WAPE)
	\begin{equation}
		\label{eqn:wape}
		\textit{WAPE} = \frac{\sum_{t=T+1}^{T+h}|e_t|}{\sum_{t=T+1}^{T+h}|y_t|}
	\end{equation}
\end{enumerate}

Similar to the modification on MAPE to obtain sMAPE, WAPE can also be modified in the denominator to form the Symmetric Weighted Absolute Percentage Error (sWAPE) measure, although this has not been defined previously in the literature, to the best of our knowledge. Similar to sMAPE, sWAPE also avoids the problem of the final error being different when $y_t$ and $\hat{y}_t$ are exchanged.

\begin{enumerate}
	\setcounter{enumi}{9}
	\item Symmetric Weighted Absolute Percentage Error (sWAPE)
	\begin{equation}
		\label{eqn:swape}
		\textit{sWAPE} = \frac{\sum_{t=T+1}^{T+h}|e_t|}{\sum_{t=T+1}^{T+h}|y_t| + |\hat{y}_t|}
	\end{equation}
\end{enumerate}
A similar version can be defined as follows, using squared errors in the numerator as opposed to absolute errors.

\begin{enumerate}
	\setcounter{enumi}{10}
	\item Weighted Root Mean Squared Percentage Error (WRMSPE)
	\begin{equation}
		\textit{WRMSPE} = \sqrt{\frac{\sum_{t=T+1}^{T+h}e^2_t}{\sum_{t=T+1}^{T+h}|y_t|}}
	\end{equation}
\end{enumerate}

WAPE and WRMSPE defined above are for a single series, whereas for multiple series, the per-series errors can be summarised using mean, median etc. In the denominator of WAPE and WRMSPE,  $T$ is the length of the training part of the series whereas $h$ is the size of the forecast horizon. 
A different version of the OOS scaling of WAPE was proposed by \citet{wong_2019} as in Equation \ref{eqn:rtae}.

\begin{enumerate}
	\setcounter{enumi}{11}
	\item Relative Total Absolute Error (RTAE)
	\begin{equation}
		\label{eqn:rtae}
		\textit{RTAE} = \frac{\frac{1}{h}\sum_{t=T+1}^{T+h}|e_t|}{max(C, \frac{1}{h}\sum_{t=T+1}^{T+h}|y_t|)}
	\end{equation}
\end{enumerate}


RTAE above is defined for a single series. In Equation \ref{eqn:rtae}, $C$ refers to a regularisation constant to ensure that the denominator does not fall below the threshold $C$. In this sense, RTAE also follows the concept of winsorising discussed above for msMAPE and thus, all the associated issues hold here as well. However, as opposed to WAPE and RTAE where the aggregation of the values in the denominator is done OOS, this aggregation can be done in-sample as well. \citet{Fotios2015} use a set of error measures in their work that follow this idea. The base error for these measures is defined as in Equation \ref{eqn:percentage_in_sample_measures} where $T$ stands again for the length of the training part of the series.

\begin{equation}
	\label{eqn:percentage_in_sample_measures}
	p^{\dagger}_t = \frac{e_t}{\frac{1}{T}\sum_{t=1}^{T}y_t}
\end{equation}

In the work by \citet{Fotios2015}, $p^{\dagger}_t$ is named as the scaled Error (sE) and ${p^{\dagger}_t}^2$ is named as the scaled Squared Error (sSE). Similarly the scaled Absolute Error (sAE) is defined in Equation \ref{eqn:percentage_in_sample_measures_absolute}.

\begin{equation}
	\label{eqn:percentage_in_sample_measures_absolute}
	p^{\ddagger}_t = \frac{|e_t|}{\frac{1}{T}\sum_{t=1}^{T}y_t}
\end{equation}

In both $p^{\dagger}_t$ and $p^{\ddagger}_t$, the error at time step $t$ in the forecast horizon is scaled by the mean of the actual values in the whole training region of the series. They are named as scaled errors to indicate similarity to the scaled errors discussed further in Section \ref{sec:scaled_measures}, where error measures are scaled by a scaling factor dependent on the time series to make them scale-free. However, in this case, since actual values of the series are used to compute the scale, they can be interpreted similar to the aforementioned percentage based measures. The error measures by using these as the base errors can be defined as below.

\begin{enumerate}
	\setcounter{enumi}{12}
	\item Scaled Mean Error (sME)
	\begin{equation}
		\textit{sME} = \frac{1}{n}\sum_{t=1}^{n}({p^{\dagger}_t}) 
	\end{equation}
	\item Scaled Mean Squared Error (sMSE)
	\begin{equation}
		\textit{sMSE} = \frac{1}{n}\sum_{t=1}^{n}({p^{\dagger}_t}^2) 
	\end{equation}
	\item Scaled Mean Absolute Error (sMAE)
	\begin{equation}
		\textit{sMAE} = \frac{1}{n}\sum_{t=1}^{n}({p^{\ddagger}_t}) 
	\end{equation}
\end{enumerate}


Another option for scaling based on actual values is to consider the OOS values (in the forecast horizon) as in the WAPE error measure mentioned above, but aggregated over all the series in the dataset. \citet{SALINAS20201181} use such error measures in their work as defined below.

\begin{enumerate}
	\setcounter{enumi}{15}
	\item Normalised Deviation (ND)
	\begin{equation}
		\textit{ND} = \frac{\sum_{t=1}^{n}|e_{t}|}{\sum_{t=1}^{n}|y_{t}|}
	\end{equation}
	\item Normalised Root Mean Squared Error (NRMSE)
	\begin{equation}
		\textit{NRMSE} = \frac{\sqrt{\frac{1}{n}\sum_{t=1}^{n}(e_{t}^2)}}{\frac{1}{n}\sum_{t=1}^{n}(|y_{t}|)}		
	\end{equation}
\end{enumerate}

\subsubsection{Measures based on Relative Errors}
\label{sec:relative_errors}
In Relative Errors, scaling is done through dividing by errors from a benchmark method. This scaling is done per each time step (the error of the model at a particular time step is divided by the error from a benchmark method such as the na\"ive or the seasonal na\"ive for the same time step). The idea is to measure the performance of the forecasting model with respect to this benchmark method. On top of being scale-independent (errors scaled with respect to the scale of the series), relative measures are useful when necessary to average over series that differ in forecastability. These measures can standardise the series for their degree of difficulty in forecasting since the models are compared against a benchmark method on the same series~\citep{Armstrong_Adya_Collopy_2001}. The relative error is defined as in Equation \ref{eqn:relative_error} where $e_t^b$ denotes the error of a benchmark method, commonly the na\"ive method.

\begin{equation}
	\label{eqn:relative_error}
	r_t = \frac{e_t}{e_t^b}
\end{equation}

Measures based on relative errors are as defined below.

\begin{enumerate}
	\item Mean Relative Absolute Error (MRAE)
	\begin{equation}
		\textit{MRAE} = \frac{1}{n}\sum_{t=1}^{n}(|r_t|)	
	\end{equation}
	\item Median Relative Absolute Error (MdRAE)
	\begin{equation}
		\textit{MdRAE} =  \textit{median}(|r_t|)
	\end{equation}
	\item Root Mean Relative Squared Errors (RMRSE)
	\begin{equation}
		\textit{RMRSE} = \sqrt{\frac{1}{n}\sum_{t=1}^{n}(r^2_t)}
	\end{equation}
	\item Geometric Mean Relative Absoluate Error (GMRAE)
	\begin{align}
		\begin{split}
			\textit{GMRAE} = \sqrt[n]{\prod_{t=1}^{n}{|r_t|}}
		\end{split}
	\end{align}
	
	Relative Geometric Root Mean Squared Error (RGRMSE)
	\begin{equation}
		\textit{RGRMSE} = \sqrt[2n]{\prod_{t=1}^{n}{r_t^2}}
	\end{equation}
	
	Although the definitions are slightly different, mathematically both GMRAE and RGRMSE are equivalent.
	
\end{enumerate}

The biggest advantage of these measures is that they are more interpretable, and directly comparable across datasets unlike those unbounded measures mentioned before. For instance, a MAPE value of 1\% in itself gives no indication whether it is a high or a low value for the particular series without any explicit benchmark comparisons. On the other hand, measures which scale based on benchmark errors give direct interpretation of how good or bad the model is with respect to the benchmark, and thus also become comparable across datasets. However, with these relative measures the relative magnitude of the overall errors that we get depends on the competence of the underlying benchmark method on the respective series. Large errors from the benchmark methods tend to lessen the impact of the errors from our models. The opposite can happen too, where the underlying benchmark method is extremely good on one of the series in the dataset (very low benchmark errors), and thus overly exaggerates the errors from the forecasting model for that particular series, compared to the others. Thus, it becomes hard to capture models that do well on such series. Therefore, the choice of the benchmark method plays an important role for relative errors.



\subsubsection{Relative Measures}
\label{sec:relative_measures}

An alternative way of estimating the accuracy of methods with respect to benchmarks is by using Relative Measures. These measures simply consider the division of an error measure for the forecasting model, by that of a benchmarking method; which resolves to the relative of error measures. For example, the relative measure with MAE can be defined as below, where $MAE_b$ indicates the MAE value of the benchmarking method for the considered period.

\begin{enumerate}
	\item Relative Mean Absolute Error (RelMAE)
	\begin{equation}
		\textit{RelMAE} =\frac{ \textit{MAE}}{ \textit{MAE}_b}
	\end{equation}
\end{enumerate}

Similar measures can be defined using the previously mentioned scale-dependent measures such as MSE, MdAE as well as scale-invariant measures such as MAPE, sMAPE etc. For example, RelMSE and RelRMSE can be defined as below. 

\begin{enumerate}
	\setcounter{enumi}{1}
	\item Relative Mean Squared Error (RelMSE)
	\begin{equation}
		\textit{RelMSE} = \frac{ \textit{MSE}}{ \textit{MSE}_b}
	\end{equation}
	\item Relative Root Mean Squared Error (RelRMSE)
	\begin{equation}
		 \textit{RelRMSE} = \sqrt{\frac{ \textit{MSE}}{ \textit{MSE}_b}}
	\end{equation}
\end{enumerate}

\citet{LaiLSTNET} introduced a different version of a relative measure defined as below.

\begin{enumerate}
	\setcounter{enumi}{3}
	\item Root Relative Squared Error (RSE)
	\begin{equation}
		 \textit{RSE} = \frac{\sqrt{\sum_{t=1}^{n}e_{t}^2}}{\sqrt{\sum_{t=1}^{n}(y_{t} -  \textit{mean}(y))^2}}
	\end{equation}
\end{enumerate}

In RSE, the root squared error of all the OOS time steps across all the series is scaled by the root squares of the difference between the actuals and the mean computed for the same data points. Therefore, the benchmark forecast used in RSE is the mean forecast by considering the OOS points across all the series. 

\citet{Davydenko2013-pa} proposed the following relative measure across series in their work, where $m$ denotes the number of series and $h_i$ the number of testing time steps in the $ith$ series. Those authors argue that the geometric mean is the more suitable operator over the arithmetic mean for summarising relative measures.

\begin{enumerate}
	\setcounter{enumi}{4}
	\item Average Relative Mean Absolute Error (AvgRelMAE)
	\begin{equation}
		 \textit{AvgRelMAE} = \left(\prod_{i=1}^{m}{\left(\frac{ \textit{MAE}^i}{ \textit{MAE}_b^i}\right)^{h_i}}\right)^{\frac{1}{\sum_{i=1}^{m}h_i}}
	\end{equation}
\end{enumerate}

Similar to relative errors, the benefit of using relative measures is that their interpretation is quite intuitive. If the value of the error measure is $< 1$, this means that the forecasts from the model are more accurate than the benchmark. If the value is $> 1$, it means the opposite that the benchmark is better than the evaluated forecasting technique. However, relative measures require more than one step forecasts from each series to compute an MAE per each series. Otherwise, in the one-step forecasts case, it basically resolves to computing a relative error per each step, which brings back all the issues of relative errors as discussed later in the next section. However, this can be resolved by computing the MAE across multiple series. Yet, scale-dependent measures such as MAE only make sense when all the series have the same scale. Other than that, problems of the relative measures depend on the pitfalls of the base error metrics chosen, as detailed in the next section. 

\subsubsection{Measures based on Scaled Errors}
\label{sec:scaled_measures}

The idea of scaled errors was first introduced by \citet{hyndman2006evaluation} as an alternative to relative errors and relative measures which compare methods with respect to a benchmark. A scaled error as discussed here, scales the error of a forecasting method by the in-sample MAE of a benchmark method such as the na\"ive method. The scaled error by using MAE for the benchmark can be defined as in Equation \ref{eqn:scaled_error}.

\begin{equation}
	\label{eqn:scaled_error}
	q_t = \frac{e_t}{\frac{1}{T-1}\sum_{t=2}^{T}|y_t - y_{t-1}|}
\end{equation}

Error measures which use the above scaled error $q_t$ as the base error can be defined as below.

\begin{enumerate}
	\item Mean Absolute Scaled Error (MASE)
	\begin{equation}
		 \textit{MASE} = \frac{1}{n}\sum_{t=1}^{n} q_t
	\end{equation}
	
	\item Median Absolute Scaled Error (MdASE)
	\begin{equation}
		 \textit{MdASE} =  \textit{median}(q_t)
	\end{equation}
\end{enumerate}

A similar scaled error by using MSE of the benchmark for the denominator can be defined as in Equation \ref{eqn:scaled_squared_error}.

\begin{equation}
	\label{eqn:scaled_squared_error}
	q_t^\dagger = \frac{e_t^2}{\frac{1}{T-1}\sum_{t=2}^{T}(y_t - y_{t-1})^2}
\end{equation}

Error measures by using $q_t^\dagger$ as the base error can be defined as below.

\begin{enumerate}
	\setcounter{enumi}{2}
	\item Root Mean Squared Scaled Error (RMSSE) - used in the M5 Forecasting Competition~\citep{makridakism5}.
	\begin{equation}
		 \textit{RMSSE} = \sqrt{\frac{1}{n}\sum_{t=1}^{n}q_t^\dagger}
	\end{equation}
\end{enumerate}

Measures based on scaled errors are symmetric, meaning that both positive and negative errors as well as errors of the same magnitude at both high-valued and low-valued points of the series, get penalised the same way~\citep{Koutsandreas2021-ww}. The interpretation of MASE is that, if the value is $< 1$, the proposed method is better than the one-step ahead na\"ive method in-sample and the opposite if the value is $> 1$. In this sense, these measures have a meaningful interpretation and according to \citet{hyndman2006evaluation}, they are applicable to a wide variety of forecasting scenarios without any problems. However, as \citet{Koutsandreas2021-ww} argue, the measure generally has limited interpretability when applied in business applications in practice. Since the comparison is performed against in-sample benchmark errors, the results are difficult to communicate. Therefore, these measures are used mostly in research related work due to their beneficial properties.

Another problem with scaling based on in-sample one-step ahead benchmark errors occurs when the evaluated model produces multi-step ahead forecasts. Using a one-step ahead in-sample na\"ive forecast as benchmark will often result in huge errors ($>> 1$) for the forecasts far ahead in the horizon, simply because the benchmark tackles an easier forecasting problem. Another problem is that this procedure makes the MASE errors coming from two datasets with two different sizes of the forecast horizon incomparable with each other. The solution is to consider in-sample multi-step ahead forecasts (with the same size as the forecast horizon) for the benchmark method as well, for reasonable comparison~\citep{hyndman2006evaluation}. However, this procedure may be complicated to implement and may hinder interpretability. Also, care needs to be taken about which benchmark is used. For certain series a one-step-ahead na\"ive may be adequate, and for other series a seasonal na\"ive benchmark. The measures cannot be compared across datasets/series when the underlying benchmark used is different in each scenario.


\subsubsection{Measures based on Ranks/Counting}

Instead of summarising base errors, we can also summarise rankings of models resulting from the base errors for the different series or different time steps in the horizon. This way we can obtain a fully scale-free error measure even by using absolute error as the base error. The M competition and the M3 competition used a ranking among the participating methods~\citep{MAKRIDAKIS2000451}. However, one problem with ranking is that it is dependent on the other competing methods. 

A similar idea to measures based on ranks is ``Percentage Better", also used in the M3 competition~\citep{MAKRIDAKIS2000451}. Here, the idea is to use a benchmark method such as random walk and count how many times (across series and time steps) a given method is better than the benchmark and report it as a percentage. In this sense, this measure also has similarities with the relative measures since the comparison is done against a benchmark~\citep{Koutsandreas2021-ww}. \citet{hyndman2006evaluation} name this measure as PB score and define it for instance with MAE base errors as follows.

\begin{equation}
	\label{eqn:pb_score}
	 \textit{PB(MAE)} = 100\  \textit{mean}(I\{ \textit{MAE}< \textit{MAE}_b\})
\end{equation}

A similar idea was proposed by \citet{wong_2019} to measure the percentage of forecasts where the value of a particular error is higher than a margin X. This error measure can capture succinctly how severe the deviations of the forecasts of the model from the actuals are and whether to take action about it. This error measure was named the `Percentage of Critical Event for Margin X' and is slightly different from Equation \ref{eqn:pb_score} as below. In Equation \ref{eqn:percentage_worse}, the error measure $E$ can be any measure defined per user requirements.

\begin{equation}
	\label{eqn:percentage_worse}
	100\  \textit{mean}(I\{E>X\})
\end{equation}

\subsubsection{Measures based on a Transformation}

There are other scale-invariant error measures which use a transformation such as logarithm on the errors which can be defined as follows. Assuming non-negative time series, to avoid problems with 0 values in the logarithm, 1 is added to both the actual values as well as the predictions. The following is defined by using the natural logarithm.

\begin{equation}
	\label{eqn:log_error}
	l_t = ln(y_t + 1) - ln(\hat{y}_t + 1)
\end{equation}

Error measures based on logarithmic errors are known to optimise for the geometric mean of the distribution~\citep{SvetunkovAdam}. The above formula is mathematically equivalent to the following Equation \ref{eqn:log_division}. Therefore, as \citet{Tofallis2015-yf} also claim this error gives rich information indicating both a difference as well as  a ratio (similar to percentage based measures). 
\begin{equation}
	\label{eqn:log_division}
	l_t = ln\left(\frac{y_t + 1}{\hat{y}_t + 1}\right)
\end{equation}

\begin{enumerate}
	\item Root Mean Squared Logarithmic Error (RMSLE) - Used for the Walmart Stormy Weather Forecasting Competition\footnote{\url{https://www.kaggle.com/c/walmart-recruiting-sales-in-stormy-weather/}} and Recruit Restaurant Visitor Forecasting Competition\footnote{\url{https://www.kaggle.com/c/recruit-restaurant-visitor-forecasting}}~\citep{BOJER2020}
	\begin{equation}
		 \textit{RMSLE} = \sqrt{\frac{1}{n}\sum_{t=1}^{n}l_t}
	\end{equation}
	
	\citet{Tofallis2015-yf} name the same error measure as Log Accuracy Ratio (LogAR). 
	\item Normalised Weighted Root Mean Squared Logarithmic Error (NWRMSLE) - Used for the Corporaci\'on Favorita Grocery Sales Forecasting Competition\footnote{\url{https://www.kaggle.com/c/favorita-grocery-sales-forecasting}}~\citep{BOJER2020}. In this particular competition, the weights for the different series were provided separately to the participants. Perishable items were assigned a weight of $1.25$ and all the others were assigned a weight of $1.00$.
	\begin{equation}  
		\textit{NWRMSLE} = \sqrt{\frac{\sum_{t=1}^{n}w_tl_t^2}{\sum_{t=1}^{n}w_t}}
	\end{equation}
\end{enumerate}

Log transformation makes the data approximately normal by scaling down the errors using a monotonic transformation. However, the transformation due to logarithm is not a bounded transformation; thus it does not scale values to a pre-specified range. Nevertheless, it has several useful mathematical properties. The RMSLE measure is symmetric meaning that interchanging $y_t$ and $\hat{y}_t$ does not change the value of the measure. Moreover, these measures produce unbiased forecasts, with the effect from the over-forecasts and under-forecasts being balanced~\citep{Tofallis2015-yf}.

\subsubsection{Other Error Measures in the Literature}
\label{sec:other_measures}

In this section, we also report a few other error measures that have been published in the literature.

\begin{figure*}[htbp!]
	\hspace{-1cm}
	\includegraphics[scale=0.9]{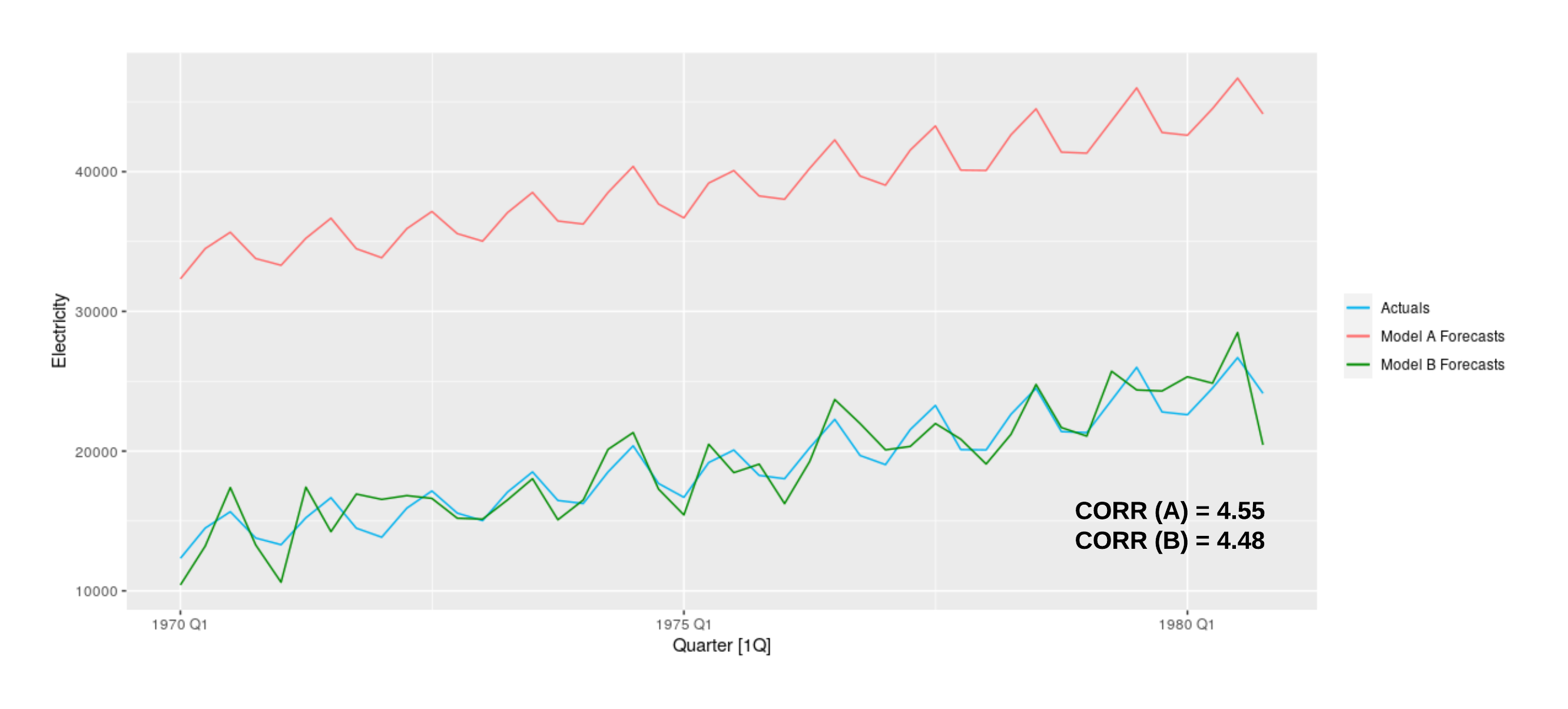}
	\caption{CORR Error Measure Issue - With Heavily Biased Forecasts}
	\label{fig:corr_issues}
\end{figure*}

\begin{enumerate}
	\item Rate-based Error Measures - Introduced by \citet{Kourentzes2014-zn} 	\begin{equation}
		\label{eqn:rate_based_error}
		c_t = \hat{y_t} - \frac{1}{t}\sum_{i=1}^{t}y_i
	\end{equation}
	Measures which use $c_t$ as the base error can be defined as below.
	\begin{itemize}
		\item Mean Squared Rate (MSR)
		\begin{equation}
			\label{eqn:msr}
			 \textit{MSR} = \sum_{t=1}^{n}c_t^2
		\end{equation}
		\item Mean Absolute Rate (MAR)
		\begin{equation}
			 \textit{MAR} = \sum_{t=1}^{n} |c_t|
		\end{equation}
	\end{itemize} 
	Measures based on rate-based errors, are in general suitable for inventory management decision making where the exact accuracy per each time step is not the interest; rather, maintaining an optimal inventory without many underestimations is more important. 
	\item Weighted Mean Absolute Error (WMAE) - Used at the Walmart Store Sales Forecasting Competition
	\footnote{\url{https://www.kaggle.com/c/walmart-recruiting-store-sales-forecasting/}}~\citep{BOJER2020}. The weights can be assigned either for particular series (when summarising across series) or particular time steps in the horizon (for promotion periods etc.) For example, in the aforementioned competition, a weight of $5$ has been assigned to the days if the corresponding week has a holiday and a weight of $1$ otherwise.
	\begin{equation}
		 \textit{WMAE} = \frac{\sum_{t=1}^{n}w_t|e_t|}{\sum_{t=1}^{n}w_t}
	\end{equation}
	With WMAE, the idea is to heighten or lessen the impact on particular series/days based on their importance by using appropriate weights. However, if the individual series have very different scales, the intended impact from a particular day from different series would still be subject to scale differences of those series. The error impact from a series with high scale will be higher than the impact on a series with lower scale (even though the used weight is the same). This can be circumvented by adjusting the weights assigned to the individual series based on their scales or by using a scale-free base error.

	\item Empirical Correlation Coefficient (CORR) - Proposed by \citet{LaiLSTNET}.
	\begin{equation}
		\label{eqn:corr}
		 \textit{CORR} = \frac{1}{m}\sum_{i=1}^{m}(\frac{\sum_{t=T+1}^{T + h}(y_{it} -  \textit{mean}(y_i))(\hat{y}_{it} -  \textit{mean}(\hat{y_i}))}{\sqrt{\sum_{t = T+1}^{T+h}(y_{it} -  \textit{mean}(y_i))^2(\hat{y}_{it} -  \textit{mean}(\hat{y}_i))^2}})
	\end{equation}	
	This definition follows the idea of the correlation coefficient to measure the linear relationship between two variables. The outer mean in Equation \ref{eqn:corr} is calculated across multiple series where $m$ denotes the number of series in the dataset. The inner mean is calculated for the different time steps of the individual series. Unlike other error measures, for CORR higher values are better. With the CORR error measure, we may select a biased model as the best model since the correlation between the actuals and the biased actuals is perfect. This scenario is illustrated in Figure \ref{fig:corr_issues} where model A is a heavily overestimating model with a better CORR value of 4.55 than model B with a lower CORR value of 4.48. Therefore, CORR will select model A as the better model although model B as shown is a better suited model for this scenario.
\end{enumerate}


\subsection{Problems of the Error Measures and Guidelines}

In this section, we outline which error measures are applicable for each of the characteristics of the underlying time series. Since there is no single error measure that is applicable universally for all circumstances, as long as we are aware of the common pitfalls of error measures and ensure that the series under consideration are free from the troublesome characteristics, it should be safe to use the respective error measures for forecast evaluation. 

Scale-dependent measures perform no scaling and are thus suitable for comparing methods across series that have similar scales, but not with different scales. When evaluating across series, the total error with scale-dependent error measures is dominated by those series that have higher scales/volumes. Because of this, a model can perform poorly on higher scaled series and really good on all the other series, but still end up as the worse model, with scale-dependent measures. However, depending on the business context, this can be a valid objective to forecast more accurately the series that have higher scales, since they may be really the objects of interest. Also, if the set of series under consideration have meaningful scales, making errors scale-free is perhaps not needed at all. This is the case with most of the real-world time series such as retail sales of certain products. However, the problem with scale-dependent measures is that, as soon as the scale of the series is changed (for example converting from one currency to another), the value of the error measures change~\citep{TASHMAN2000437}. 
We may sometimes be even more interested in the final dollar value of the sales rather than the actual sales volume itself. In such a scenario, the errors need to be calculated in the respective scales of interest. Furthermore, if the underlying series/datasets contain no scale differences at all, the best option to compute errors is by using scale-dependent RMSE/MAE error measures which are robust against scaling in many practical time series scenarios. The errors computed likewise can be used in statistical tests for significance of the differences as described in Section \ref{sec:statistical_tests} to get a better picture of the competitiveness of the used models.

The need for scale-free measures comes from the wish to be able to evaluate models across time steps/series/datasets, with different scales. For example, if we say that MAE is 10 for a particular series, we have no idea whether it is a good or a bad accuracy. For a series with an average value of 1000, this amount of accuracy is presumably quite good, whereas for another series with an average value of 1, it is a very bad accuracy. If we want to obtain models that in general produce good forecasts for all the time series despite their scales, then scale-invariant measures are the better option. The underlying objective of making errors scale-free is that all the errors (coming from different time steps/series/datasets) contribute to comparable amounts of influence in the final overall error measure value, instead of some errors always dominating over others due to their scales. The error measures seen in the literature, have different forms of scaling; per-step, per-series or per-dataset. The particular scaling and therewith the error measures selected, depend on the underlying intentions of practitioners. In general, applications which intend to have comparability across differently scaled time steps of a single series may be more interested in per time step scales. Per series scaling holds for comparing among differently scaled series (with respect to forecastability or volume) within a dataset, but not necessarily for such differently scaled time steps within an individual series. On the other hand, global scales apply mostly for similar scaled time series within datasets, but have totally varying scales on different datasets, which we want to compare the models across. This is because, within a dataset, a global scaling does not perform a scaling per each series, but rather scaling all of them, due to simply dividing by a constant for the dataset, which still makes the errors scale-dependent in their relative differences. Therefore, if we are confident that the series within a single dataset are similar scaled, this type of scaling can be used to select models that perform well in general across different datasets with different scales, or to investigate how the performance of models differ across datasets by comparing. Log transformed errors also perform a scaling down of all the errors. However, log transformation is a monotonic transformation and therefore, unlike global scaling, log transformed errors are not quite suitable for comparison across datasets having different scales. 



Apart from that, counting/ranking is a good option for making errors scale-free in general, since there are no issues associated with the computation of the scale arising due to non-stationarities of the time series considered. Scale-dependent measures such as RMSE, MAE also avoid issues arising from the computation of the scale in scale-free measures (as further
detailed in the following). Therefore, if comparison across different scaled series is a necessity, a more robust way of making errors scale-free is to compute RMSE/MAE values and consider the ranks of the involved models per series according to those measures. Ranking ignores the actual size of the errors, this can effectively avoid the errors being biased towards higher scaled series. However, due to the same reason, one method can perform marginally better than all the other methods in all the series except one where it is drastically worse than all the others. In such a scenario, a measure based on ranking would select that method as the best whereas in reality we would be more interested in a method that performs generally good on all the series. PB score too follows the same behaviour.

Traditionally, in forecasting, the scale of series was only a problem at the end, because methods were trained per series but evaluated across series. Now, with global/pooled models which learn across many series, we have these scaling problems already at the model building phase, i.e.\ when normalising the input data. If the series have meaningful scales, and the final evaluation is expected to be performed on these original scales, it does not make sense to perform the modelling initially on pre-processed scale-free series. On the other hand, as is the practice in the ML community, most ML based models (apart from tree based models) are known to perform better with normalised data where all data points have similar scales. This makes the convergence of these models to optimal parameters much faster and accurate. Therefore, this poses a form of a contradiction between the model building and model evaluation phases for a global model. This can be addressed to a certain extent by using other techniques such as enforcing a partial scale dependence in the loss function used for model training or use the scale normalisation only for feeding inputs to the models and compute the losses on the original scale etc. As we deem the pre-processing of data to be outside the scope of our work, we focus in the following on the error measures and problems. However, very similar considerations hold for normalising data as pre-processing for ML methods.  

Apart from that, from a business perspective, we are more interested in calculating different error measures and estimating the model performance from many dimensions/aspects collectively~\citep{google_session_2020}. Evaluating the same forecasts with respect to many evaluation measures is a form of sanity checking to ensure that even under other measures (though not directly optimising for them), the forecasts still perform well. Furthermore, within a more practical context, forecasting and the respective accuracies of models alone are not sufficient, but rather how the produced forecasts affect the downstream decision making processes. In this sense, it can be argued that the final accuracy of the evaluated models need to be closely tied to business utility as seen by the downstream processes being optimised, such as energy costs, storage costs and trade-offs between them etc. This effectively makes the loss functions of models dependent upon each application specific requirement by having different constraints imposed. For example, with respect to short-term air quality prediction, it is more important to predict the peaks (extreme events) right, to detect potential health hazards. Similarly, in terms of wind power prediction, ramp forecasting (forecasting large and rapid variations in wind power) is of more importance, since that may help better plan for costs of extreme power integration into the grid and other power system operations. Hence, in both these scenarios probabilistic forecasting may be more interesting to understand how reliable the extreme forecasts would be. 

In the work by \citet{SPILIOTIS2021108237}, using the M5 Competition dataset they demonstrate that simpler empirical methods may outperform sophisticated ML models when the focus is on the downstream inventory optimisation processes. Therefore, oftentimes, the relationship between accuracy and final business utility is not a linear one indicating that a fine-grained improvement of one model over another in terms of forecasting does not necessarily guarantee significantly improved business value. \citet{abolghasemi2021stateoftheart} have also emphasised the same with respect to their first place solution at the IEEE-CIS Technical Challenge on Predict + Optimize for Renewable Energy Scheduling~\citep{ieee-cis}. Forecasting and optimisation are considered as difficult problems by themselves separately. While methods can be developed by combining the two together, this inevitably results in quite complex loss functions customised for each specific business scenario. Although such exhaustive discussions would be useful, we opt to limit the scope of this work to analyse how the evaluation measures are influenced by other common patterns and characteristics of time series, due to the lack of practicality in the former approach.

In the following, we conduct an in-depth analysis on which error measures out of those mentioned in Section \ref{sec:error_measures_categorisation} are applicable under different characteristics of the underlying time series.



\subsubsection{Count Data Well above Zero with Stationarity}

\begin{figure*}[htbp!]
	\centering
	\includegraphics[scale=0.67]{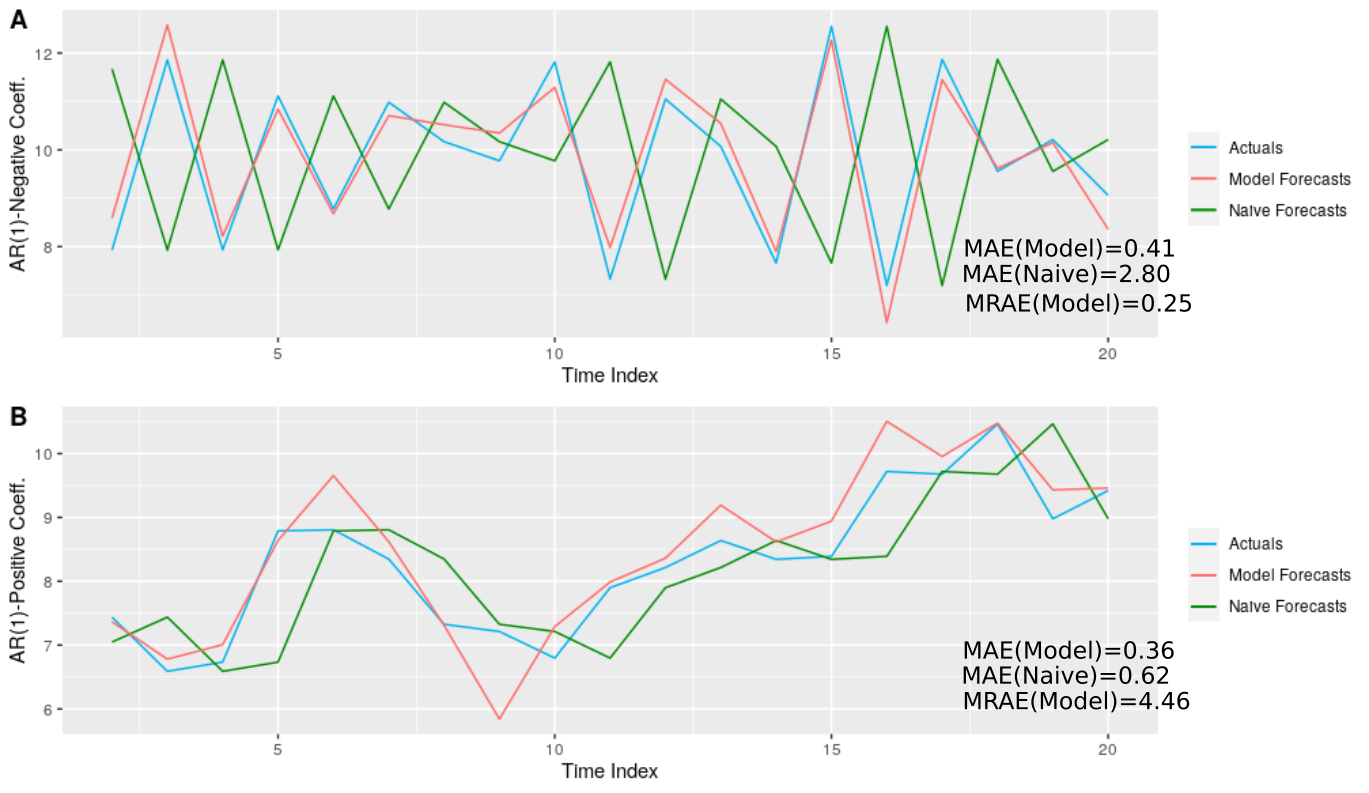}
	\caption{Relative Errors Issue - On a Series Following an AR Process with a Negative Coefficient)}
	\label{fig:relative_errors_issues_3}
\end{figure*}

Under this category we imply series having no special non-stationarities such as trends or seasonalities. Apart from these characteristics, these types of series are also bounded by certain upper and lower limits. For series like these having no problematic characteristics generally any error measure is applicable. However, on certain types of series such as an AR process with a large negative coefficient at a low order lag where the series keeps fluctuating considerably for each time step, measures which scale based on errors from a na\"ive benchmark method, need to be used with caution. This is because on such series, the na\"ive method will usually have bad performance, even when considering rolling origin to constitute the forecast horizon. Figure \ref{fig:relative_errors_issues_3} depicts this scenario for the rolling origin na\"ive forecast. According to this figure, the model forecast in scenario $A$, which has a slightly higher MAE of 0.41 than scenario $B$, has resulted in a much lower MRAE of 0.25 than the 4.46 in scenario $B$, simply due to the bad performance of the na\"ive forecast in scenario $A$.

Furthermore, relative measures face similar issues of relative errors where the errors depend on the relative competence of the benchmark method in the intended forecast horizon.

\subsubsection{Seasonality}

\begin{figure*}[htbp!]
	\centering
	\includegraphics[scale=0.58]{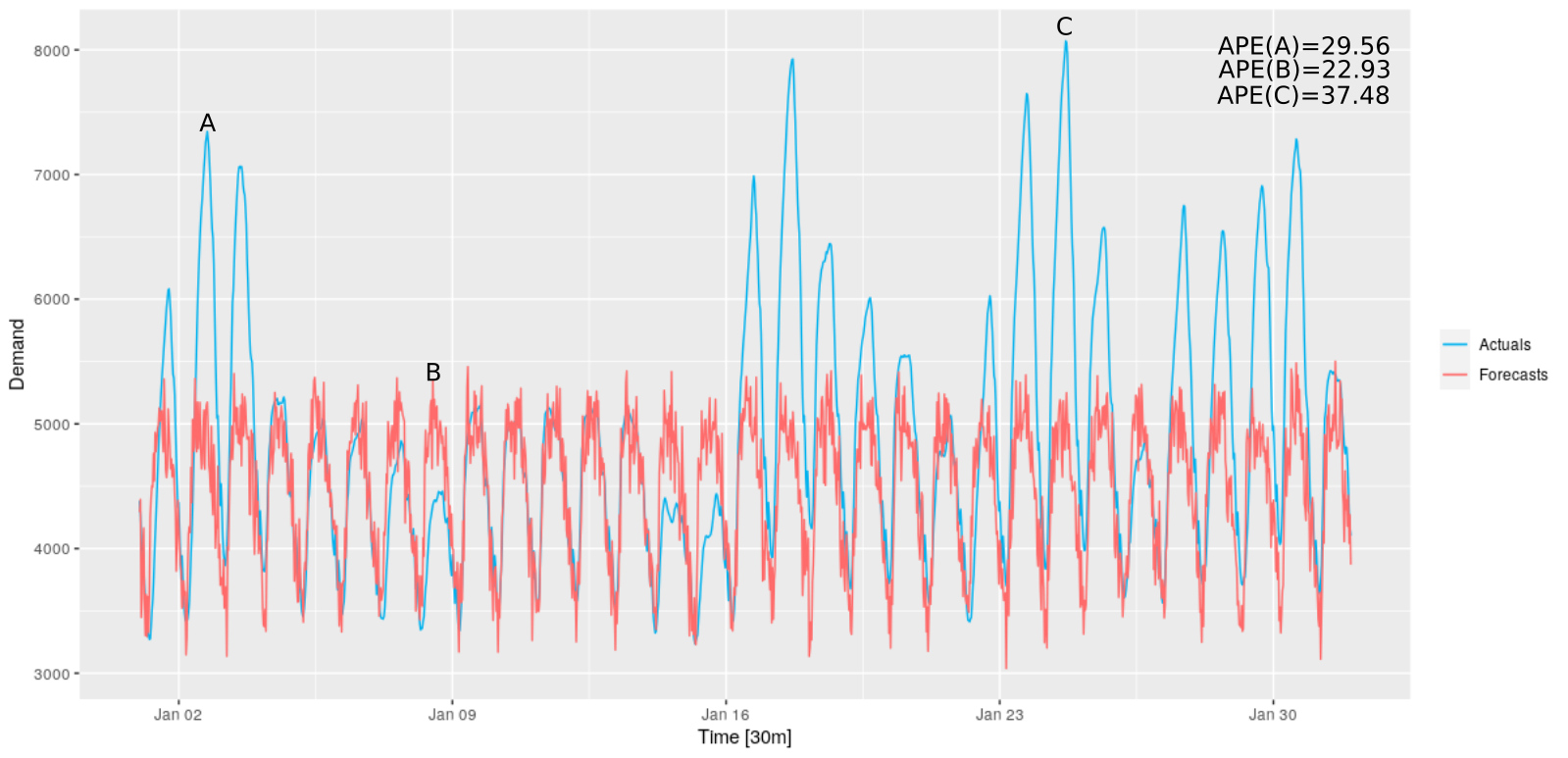}
	\caption{MAPE Error Measure Issue - On Series with Seasonality}
	\label{fig:mape_issues_1}
\end{figure*}

On series having seasonality, percentage based measures tend to underestimate the errors at uncaptured peaks of the time series heavily, due to dividing by large actual values in the percentage~\citep{wong_2019, kunst}. On the other hand, when the actual values of the time series are quite small in magnitude, percentage based measures overstate the errors due to noise. This scenario is depicted in Figure \ref{fig:mape_issues_1}, where for the three points $A, B$ and $C$, Absolute Percentage Error (APE), is indicated separately. According to this scenario, the large error at peaks $A$ and $C$ get more or less similar attention to the small error at the relatively lower-scaled point $B$. Averaging to compute MAPE, further decreases the effects on uncaptured peaks on this series. Depending on the business scenario, predicting these peaks accurately may be important, rather than focussing on the smaller fluctuations due to noise; for example, the peak demands of electricity. Therefore, percentage based measures such as MAPE, MdAPE, RMSPE can be problematic with seasonal series.

However, this problem of percentage based measures on seasonal series can be largely overcome by using measures such as WAPE or ND, NRMSE which compute the scale based on several aggregated time steps or series. 
When scaling based on benchmark errors is used, on series having seasonality, it needs to be ensured that the used benchmark method can capture seasonality, such as a seasonal na\"ive method, to be competitive against the evaluated forecasting method. 



\subsubsection{Trends}
\label{sec:error_measures_with_trends}
\begin{figure*}[htbp!]
	\hspace{-1.4cm}
	\includegraphics[scale=0.36]{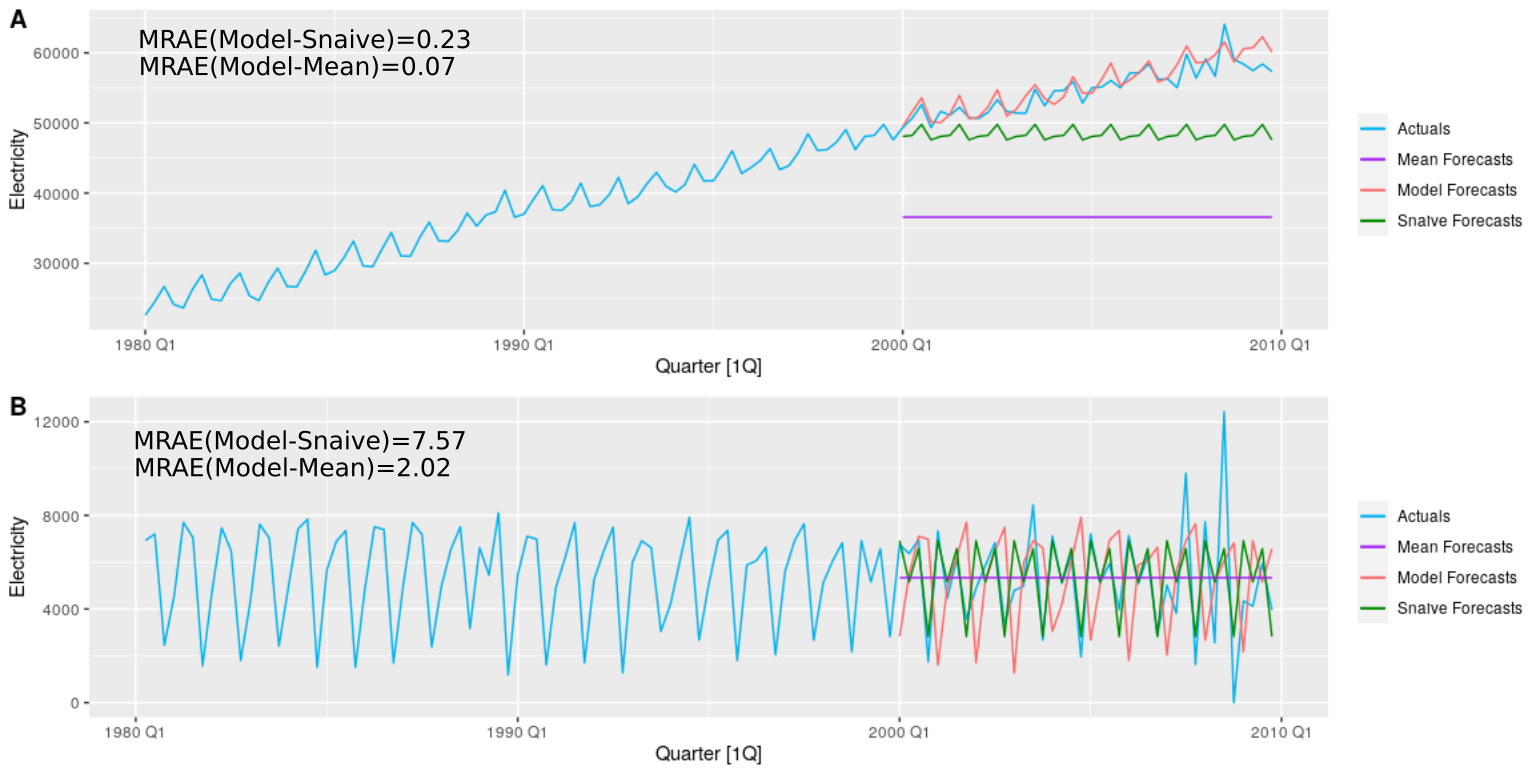}
	\caption{Relative Errors Issue - On Trended Series}
	\label{fig:relative_errors_issues_2}
\end{figure*}

When there are strong trends existing in the series, scale-free measures which compute their scale by aggregating the values (actual values or benchmark errors) at several time steps, tend to face problems. This situation is in resonance with the phenomenon explained by \cite{chen_twycross_garibaldi_2017} that the error values at each time step need to comply with the scale of the series at each point. A scale computed by aggregating over several time steps which include non-stationarities/non-normalities such as strong trends may not always be a good estimator to represent the scaling factors for all the time steps of such a series. For example, on a series having a strong upward trend, when calculating WAPE as the error measure, a certain amount of discrepancy between the actuals and the forecasts can get a lower attention than the same amount of discrepancy on a series having no such strong trends. The opposite can happen too, due to downward trends. The same situation occurs with in-sample values based scaling such as sMAE, sMSE. With Relative Errors and Relative Measures as well, on series having strong trends/heteroscedasticity, if the benchmark is fixed origin mean forecast or the na\"ive method, the benchmark errors become large on such non-stationary series as depicted in Figure \ref{fig:relative_errors_issues_2}. This results in very small overall relative errors for the evaluated model, as in scenario $A$, although the model here seems to have similar performance to the model in scenario $B$. 

To deal with this problem related to non-stationarities of series when computing the scale of percentage based measures, some form of rolling-window based summary statistic computed along the series can be used. The size of the windows are to be decided to fit the context of the underlying series. However, for measures such as MASE, RMSSE, which scale based on in-sample benchmark errors, this is not a problem. MASE can compute comparable amounts of scaling for both stationary and non-stationary series with for example trends, since the in-sample error from the na\"ive method is considered which can account for trends in the training region. 

\begin{figure*}[htbp!]
	\includegraphics[scale=1.7]{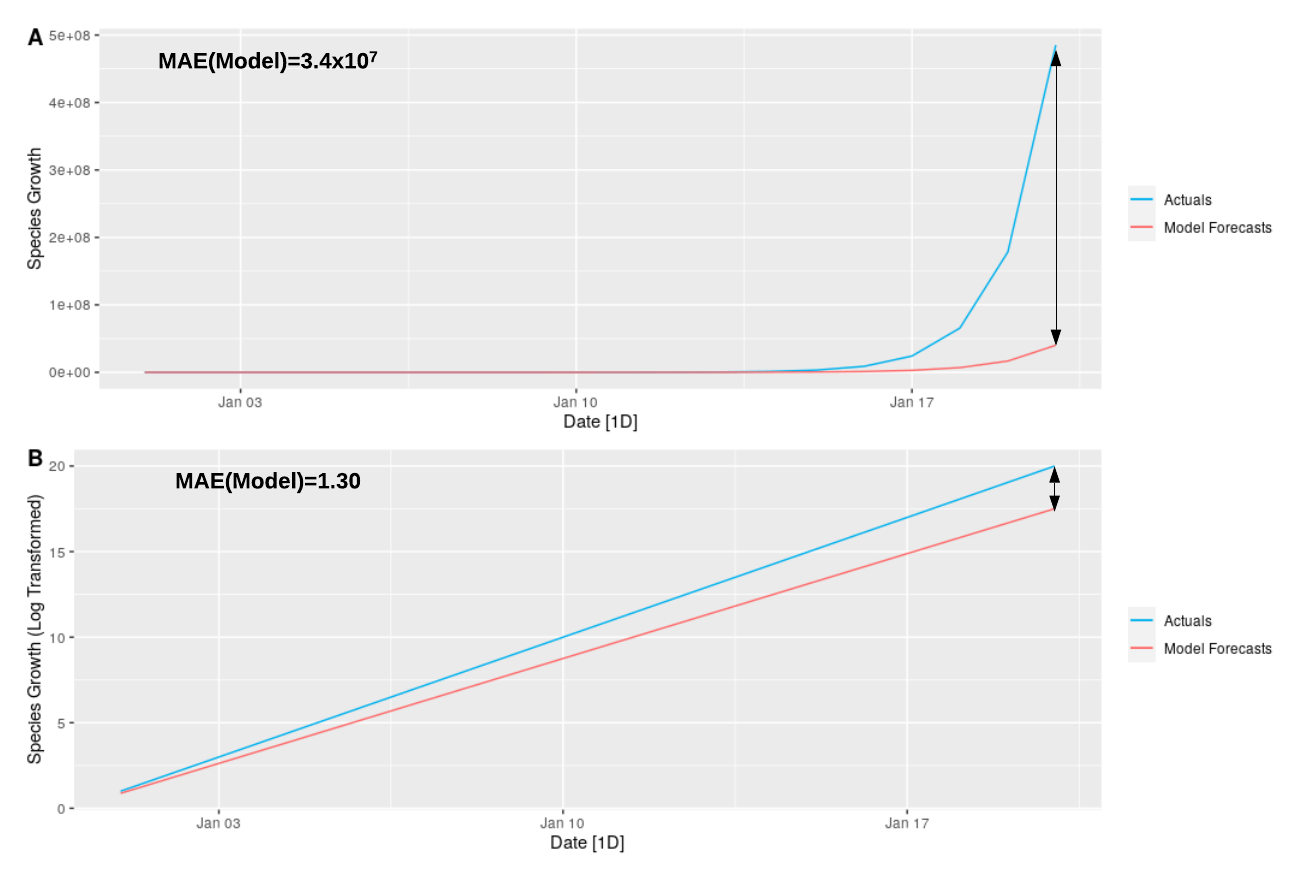}
	\caption{Transformation based Measures Issue - On a Series having Exponential Trend}
	\label{fig:log_errors_2}
\end{figure*}

Apart from that, when a global scale is computed, as with the ND measure, it is no different from measures such as RMSE, MAE which are scale-dependent. Because, the scaling factor that they compute is constant for all the time steps across all the series. Therefore, they are clearly not used in place of measures such as MAPE on trended series, where non-stationarities are accounted for, in the scaling factor per each time step. It is also important to be aware that, on series especially having exponential trends, when using log transformation based error measures, they greatly reduce the impact of errors from models. This is indicated in Figure \ref{fig:log_errors_2}.

\subsubsection{Unit Roots}

The characteristics of a series resulting from unit roots are very similar to trends. They have stochastic trends, where the direction of the series at each time step is random (due to noise) in contrast to deterministic trends. However, on a series like this, the na\"ive forecast is a quite competitive benchmark and as explained in Section \ref{sec:benchmarks}, and it is essential to compare against it. Scaled Errors such as MASE provide this support directly. On the other hand, since level changes are often seen with these unit root based series, the error measures which are suitable or unsuitable here are the same as with trended series explained in the previous section. This means that measures which compute their scale based on aggregate in-sample or OOS values or Relative Errors may tend to have problems on unit root based series. However, measures which compute a per-step scaling based on actual values, such as MAPE, RMSPE may also have issues with these series not capturing peak points similar to seasonal series.

%

\subsubsection{Heteroscedasticity}
Heteroscedastic series experience a change in the variance of the data over time. This happens mostly due to noise embedded in the time series data. With respect to forecast evaluation, this is not a very problematic characteristic since it is not associated with level changes in the time series. However, due to potential peaks and troughs in the series which may have very high and low variances due to the heteroscedasticity, measures such as MAPE and RMSPE may have problems with capturing those peaks similar to seasonal time series. Apart from that, log transformation based errors can reduce the impact from heteroscedasticity, consequently also suppressing the errors of models at such points with high variance in the data.

\subsubsection{Structural Breaks (with Level Shifts)}

\begin{figure*}[htbp!]
	\includegraphics[scale=1.4]{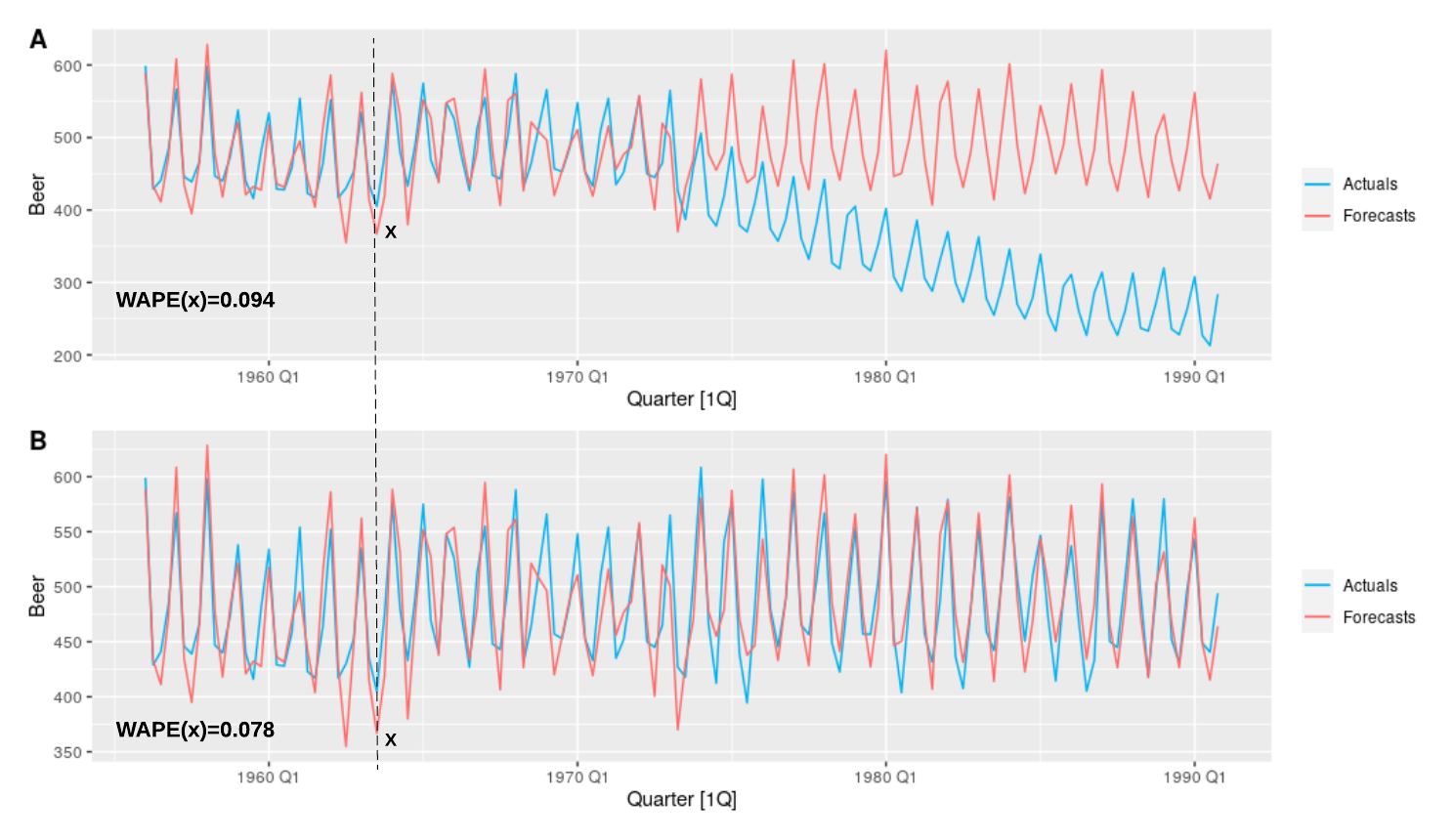}
	\caption{WAPE Error Issue - On Series with Structural Break in the Horizon}
	\label{fig:wape_issues_1}
\end{figure*}

Due to trend changepoints/structural breaks existing in the horizon, especially if we are predicting over a long horizon, WAPE can have problems as with trends in Section \ref{sec:error_measures_with_trends}. This scenario is illustrated in Figure \ref{fig:wape_issues_1} where there are two regimes within the forecast horizon with a changepoint in scenario $A$. Regime change here is the same as concept drift that we refer to in ML. Because of this non-stationarity, the scale/mean of the series is not consistent throughout the series. Due to the downward trend in the second regime, the overall scale of this series becomes smaller than in series $B$ with no regime shifts. Thus, the same amount of error (with same forecast and same actual value) at point $X$, gets a higher attention on series $A$ than on series $B$.

\begin{figure*}[htbp!]
	\includegraphics[scale=1.6]{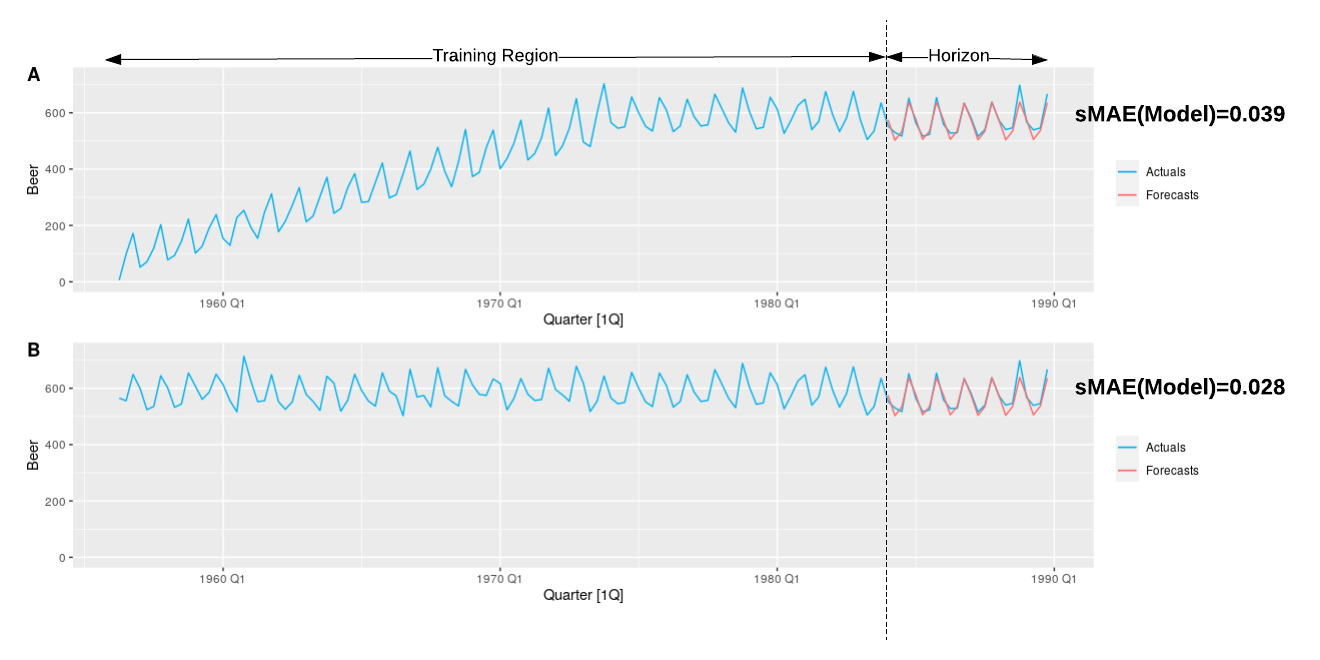}
	\caption{In-sample Scaling based Error Measures Issue - On a Series with Structural Break In-sample}
	\label{fig:in_sample_scaling_issues}
\end{figure*}

A similar situation applies to error measures which scale based on in-sample values, such as sMSE, sMAE. Due to structural breaks (anywhere in the series including in-sample, forecast horizon or the forecast origin), the mean of the series is not expected to hold constant throughout the whole series. For example, the same amount of $e_t$ in the horizons of two series, with same actuals and same forecasts from the model, can result in different overall errors when the values in the training region of the two series are different from each other, due to for example structural breaks in-sample.
This scenario is depicted in Figure \ref{fig:in_sample_scaling_issues}, where the series $A$ has a structural break in the training region, but has the same actual values as series $B$ in the forecast horizon. The model too has produced the same forecasts for both series $A$ and $B$, in the forecast horizon. This means that $e_t$ is the same for both series. However, due to scaling based on in-sample values, for series $A$, sMAE gives a slightly higher value than on series $B$. With OOS global scaling, the situation is similar to trends. On a series having non-stationarities such as a structural break, such measures are not used in the idea of scaling with respect to each time step or series, since the global estimator if the scale is a constant one.


\begin{figure*}[htbp!]
	\includegraphics[scale=1.6]{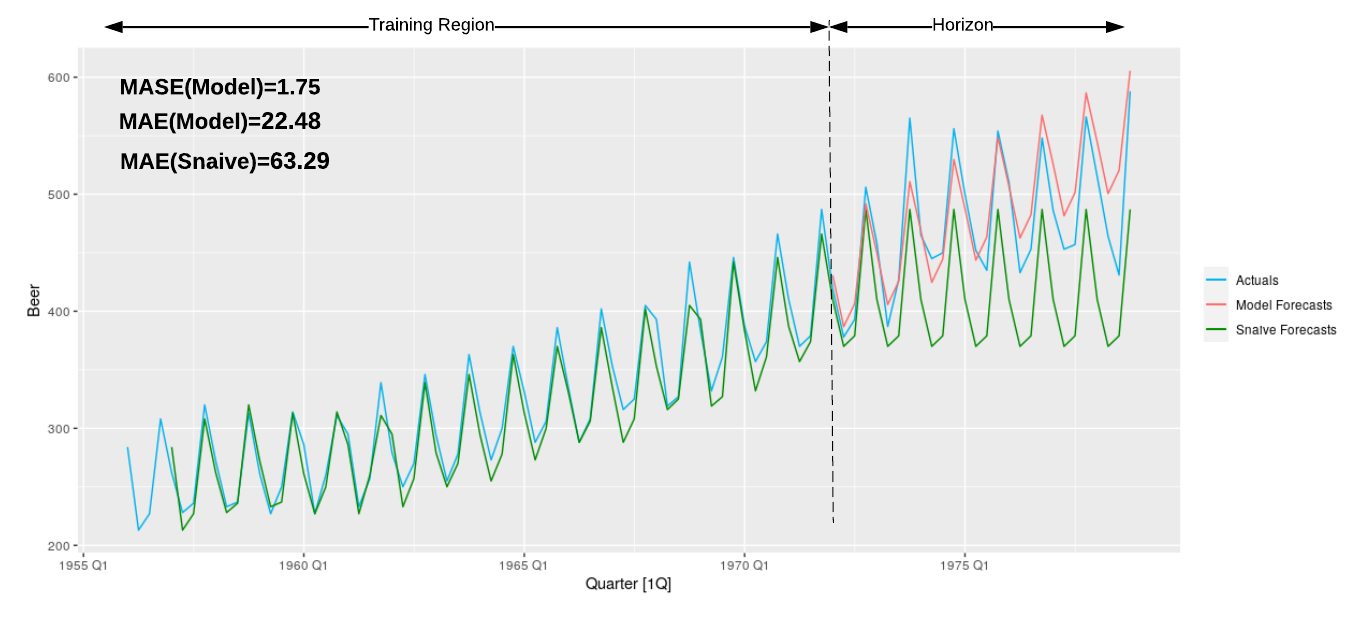}
	\caption{Scaled Errors Issue - On series with a Structural Break}
	\label{fig:mase_issues_2}
\end{figure*}

On the other hand, measures which compare against a benchmark are applicable as long as the used benchmark is comparable in accuracy to the model in the intended forecast horizon. This holds true for relative errors and measures (purely based on OOS errors), on series with structural breaks anywhere in the series. The evaluated model is expected to perform poorly on such a series when for example the structural break is in the horizon or the forecast origin. The same applies for the benchmark OOS as well. However, if the mean forecast is used as the benchmark instead of the na\"ive method, in the particular case when there are structural breaks in the training region, the evaluated model may be quite good OOS, but the mean forecast both in-sample and OOS may be badly affected by the changepoint. This is because, the mean forecast depends on the whole history of the series, including both before and after the changepoint. On scaled errors too, potential structural breaks in the horizon or the forecast origin can become problematic. The in-sample na\"ive will perform well whereas the OOS forecast from the proposed method may be completely different from the actuals. This situation is illustrated in Figure \ref{fig:mase_issues_2} where a changepoint existing at the forecast origin makes the training region different from the testing region. Hence, considering in-sample errors of the benchmark for the scaling factor in this case has resulted in a MASE value higher than 1, which means that the model is worse than the seasonal na\"ive benchmark. However, OOS it is seen that the benchmark method has even a higher MAE than the model due to the structural break. Therefore, high or low value of this error measure is not necessarily equivalent to good or bad performance. It needs to be ensured that the scale computed in-sample with benchmark errors is a reasonable scale for the errors OOS. 

\subsubsection{Intermittent Series}

With intermittent series where the distribution is heavily skewed with mostly 0's (e.g., common in retail), the mean of the series is typically higher than 0 and the median is a 0 if more than 50\% of the values in the series are zeros. However, on such intermittent series, the non-zero values are important and need to be captured well. Therefore, on such series, RMSE which minimises for the mean will be a better option to select models that predict non-zero values while MAE may select models that predict constant 0's, or are at least heavily biased towards zeros. Due to the same reason, when used as a loss function for model training too, RMSE is better on intermittent series than MAE. Therefore, using squared errors with mean operator for aggregating the errors is usually the better option for this type of series.

With any percentage based error measure, $p_t$ can be undefined when the actual values $y_t$ are 0 or close to 0, which is a very common phenomenon with intermittent time series observed in the real-world. When the actual values are close to 0, the error distribution becomes heavily skewed with extremely large errors. But, these large errors do not mean that the model performs poorly; they are simply a result of low actual values of the series~\citep{Davydenko2013-pa}. Those large error values are produced from the error measure despite the forecasts from the model at those points. It may be the case that the model produces a very accurate forecast quite close to 0, but the division by the actual value results in a very high error. On the other hand, even with an inaccurate forecast at such a point, the final error would be equally very high. Due to this reason, measures such as MAPE are not competent in distinguishing good and bad forecasts at actual values of 0. When the actual value and the forecast are both 0 too (perfect prediction at 0
actual values), MAPE becomes undefined. Some software systems that deal with MAPE error values simply disregard the errors at 0 actuals when computing the overall error measure. But, this has teh problem that it means that the accuracy of forecasts for 0 actuals are not important, which depending on the application may be true or not.~\citep{kolassa_2017}. It also means that such a modified measure does not estimate the mean nor the median nor any other meaningful summary statistic of the distribution that was originally intended to be optimised for. Using  MdAPE on intermittent series instead of MAPE also results in a slightly more robust error measure to 0 actual values. However, this also makes MdAPE less sensitive to errors at the 0 actuals, which are important and meaningful, and may overlook problems of the underlying model at such points~\citep{Davydenko2013-pa}.

The problem of division by values close to 0 in MAPE is addressed to a certain extent in sMAPE. Yet, if $y_t$ is close to 0, it is highly probable that $\hat{y}_t$ is also close to 0. 
So the problem of undefined values arising from division by values close to 0 still exists in sMAPE. Furthermore, if the actual value is 0, regardless of the value of the forecast (unless equal to 0), the sMAPE value ends up as 200 (the maximal value)~\citep{SYNTETOS200636} even though the difference between the actual and predicted values can be quite small. Therefore, in the case of intermittent series, if sMAPE is used as the evaluation measure, we effectively require the underlying forecasting technique to predict actual zeros, for example through a zero-inflated model, since the error measure does not capture good forecasts at such points, similar to MAPE. 
The msMAPE with its modified denominator, specifically addresses the issue of division by values close to 0 in sMAPE. It provides a very straightforward fix/adjustment for the issue of division by 0 values, but it has issues as discussed in Section \ref{sec:percentage_measures}. In the MAAPE measure, The $arctan$ function is well defined for all real values. Thus, unlike $p_t$, the base error in MAAPE approaches $\pi/2$ when division by 0 occurs. However, when both $y_t$ and $\hat{y}_t$ are 0 (perfect prediction), MAAPE becomes undefined. Therefore, predicting 0's right in a series can break this measure similar to sMAPE and MAPE. On the other hand, for any forecast other than 0, at a time step of an actual 0, MAAPE produces its maximum value of $\pi/2$, despite how close the forecast is to 0. In this sense, MAAPE shows a similar behaviour to sMAPE and MAPE where for actual 0's, the error measure is not consistent.

Another technique for avoiding division by 0 issues in error measures is to consider multiple time steps in the denominator as opposed to just one. This is what measures such as WAPE try to achieve. The actual values of the series for the whole forecast horizon are summed up before the division. This greatly reduces the risk of dividing by 0 values in the original percentage error $p_t$, where the division happens per each time step. WAPE is undefined only when the total of all the actual values corresponding to the forecast horizon are zero. However, that too is not that rare with a short forecasting horizon and very intermittent series. Nevertheless, due to this robustness, WAPE is generally more preferred in application scenarios than MAPE. However, due to using absolute errors, WAPE too has the problem that, on intermittent series it may select models predicting constant 0's. This can be circumvented by including squared errors $e^2_t$ in the numerator as in the WRMSPE measure. The RTAE measure also tries to fix the problem of WAPE when having all 0's or close to 0's in the whole forecast horizon. As discussed in Section \ref{sec:percentage_measures}, this is done by using a fix similar to msMAPE which has its own issues. On the other hand, with percentage based measures that scale based on in-sample values, on short intermittent series, the scale can still be 0, even when computed by aggregating several values, which is rare but possible. Specifically with the sMAE measure, it has the same problem that due to computing absolute errors, an all 0's prediction can be selected as the best on an intermittent series. This is also the case with the ND measure. However, the NRMSE measure overcomes the problem due to squared errors. Apart from that, global scaling based measures are oftentimes good measures on intermittent series, since the global summary statistic computed for the scale is less susceptible to failures on a set of intermittent series. 

With respect to relative errors, due to using absolute errors, MRAE, MdRAE and GMRAE also have the problem with intermittent series that, constant 0's can be selected as the best forecasts. Using squared errors $r_t^2$, as in RMRSE overcomes this. However, the error of the benchmark method can be very small on intermittent series. For example, if the actual value at the forecast origin was 0, the na\"ive forecasts for the whole forecast horizon would be all 0's, which results in an exact match at a 0 actual in the horizon of this intermittent series. This results in division by values close to 0 which makes relative errors undefined~\citep{hyndman2006evaluation}. The same issue is encountered, if both the benchmark and the forecasting model produce 0 errors. The solution of winsorising errors proposed by \citet{ARMSTRONG199269} in this context has the same issues as discussed for msMAPE in Section \ref{sec:percentage_measures}. 
Another solution proposed by \cite{KOLASSA2016788} for this is to sum the benchmark errors over multiple time steps and then take the ratio. This idea follows the concept behind WAPE defined in Section \ref{sec:percentage_measures} and is similar to the Relative Measures as discussed in Section \ref{sec:relative_measures}. In this case, such undefined values can only occur if the benchmark error is zero for all the considered time steps, which is a very rare case. However, this too is likely with a very short forecast horizon on an intermittent series.

Similar to the comparison between MAE and RMSE, RMSSE is better suited than MASE, with intermittent series, for the purpose of capturing spikes well. Apart from that, if all the historical observations are equal or 0 (rare but possible with a short intermittent series) the MASE and RMSSE can be infinite/undefined.

The rate-based measures mentioned in Section \ref{sec:other_measures} are designed specifically for the purpose of intermittent demand forecasting, in an inventory management context. The usual scale-dependent measures such as MSE and MAE have been criticised in this context, since they may bias the forecasts towards zero demand due to the high sparsity of the series and consequently disrupt inventory management for intermittent demand. Moreover, techniques such as Croston's method~\citep[see, e.g., ][]{robgeorg2018otext} which are specifically developed for intermittent series, forecast a demand rate (average expected demand in each period) as opposed to an exact demand for each time step, for the convenience of making inventory management decisions. Therefore, rather than comparing the demand and forecast per each time step, a rate-based error has been proposed to compare the cumulative mean of the actual demand computed over time to the intermittent demand forecasts produced by Croston-type models.

\subsubsection{Outliers}

Susceptibility to outliers depends on the underlying business needs. In some applications we may be interested in capturing outliers well (e.g., in intermittent series), whereas some others may require the models to be robust against them. The choice between squared or absolute errors and the operator used for aggregating errors mostly depend on such considerations. To be robust against outliers, a summary operator other than mean (such as median or another quantile) can be used. Geometric mean is also a generally robust option for summarising errors in the presence of outliers. Using the geometric mean has been recommended by \citet{Davydenko2013-pa} especially since it takes into account all values even in the tails of the distribution to compute the final summary, as opposed to the median. Moreover, anomaly detection techniques can be applied to identify and remove such outliers from series (relevant threshold values for the scale can be used), if they are unimportant~\citep{Arnottjfds.2019.1.064}. 

With scale-dependent measures, a model which performs generally well can end up being the worst due to large errors at outlier series. Due to considering the square of the error, RMSE and MSE are both more susceptible to outliers than MAE and MdAE which use absolute errors instead of squared errors~\citep{hyndman2006evaluation}. Similarly, measures such as RMdSE are also better with outliers due to using the median operator for summarising the errors as opposed to mean which is affected by outliers. Therefore, if the outliers are of interest and capturing them is important, using squared errors with mean operator for aggregating the errors is the better option.

\begin{figure*}[htbp!]
	\includegraphics[scale=1.3]{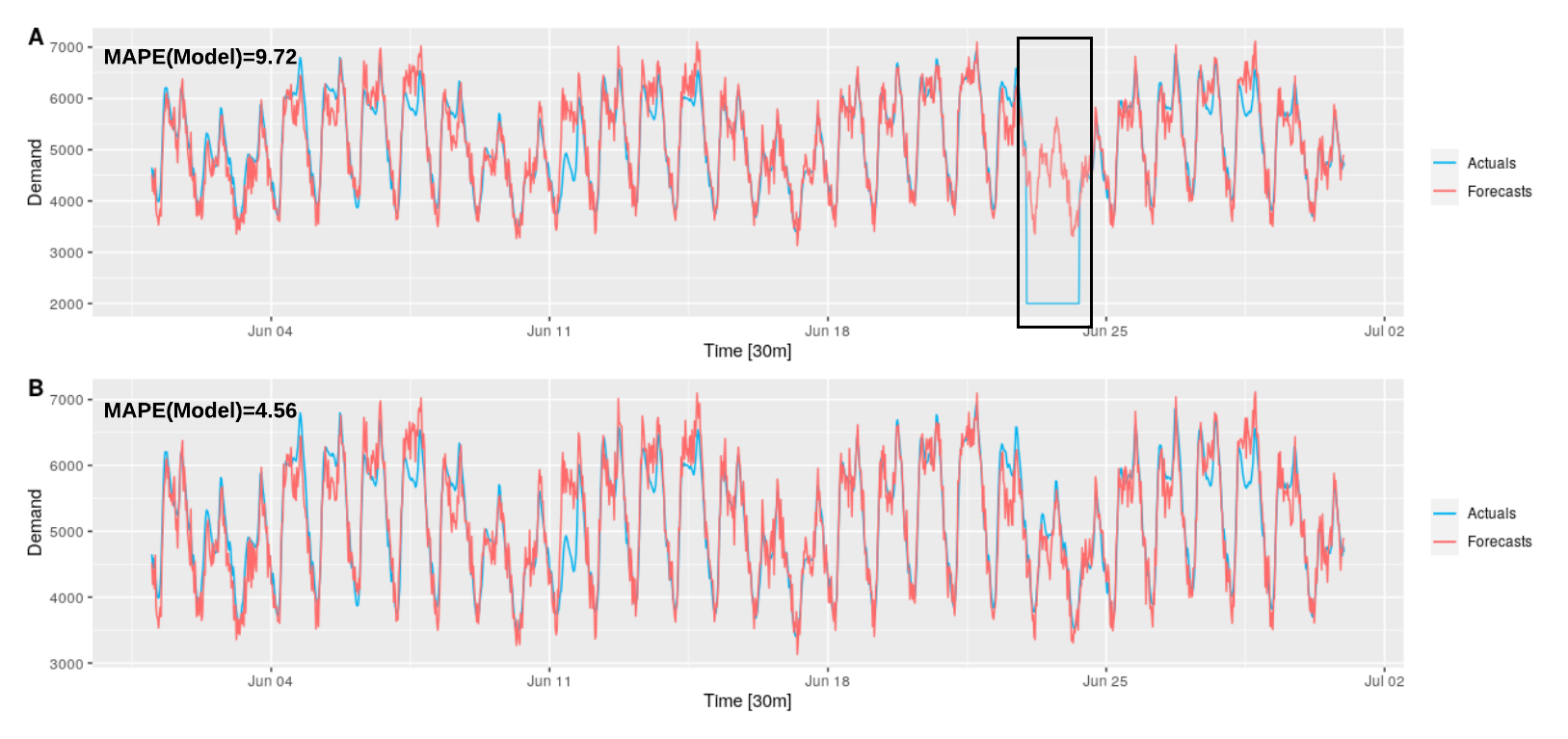}
	\caption{MAPE Error Measure Issue - On a Series having a Low Valued Outlier}
	\label{fig:mape_issues_2}
\end{figure*}

MAPE is less sensitive to errors at higher valued outlier points in the test region. On the other hand, at outliers with unexpectedly low values, the overall error gets dominated by the errors at such points. This is illustrated in Figure \ref{fig:mape_issues_2}, where series $A$ and $B$ are the same except for a low valued outlier in series $A$. The model forecasts on the two series are also exactly the same. However, because of the outlier in series $A$, the overall MAPE of the model on series $A$ is higher than on series $B$. MdAPE can be robust against outliers.
sMAPE has the advantage that it is bounded by 200, since the denominator is never less than the numerator. This makes it robust to outliers unlike other measures which are unbounded~\citep{chen_twycross_garibaldi_2017}.


The problems with extreme values described above for the MAPE are possible with WAPE too. Even if a mean on the actual values over the horizon is computed for the scale, the impact from such extreme values can be large. The problem is even worse here than with MAPE, since that same scale is used for the errors of the whole horizon. Thus, the problems at a single time step in the forecast horizon, can affect the errors from the whole horizon. For example, if there exists at least one outlier with an extremely large value in the test region, the denominator becomes very large and totally diminishes the effects from all the other errors for that particular series. The opposite can happen too, where due to very low valued outliers, even small errors from the forecasts of the series can have a large impact in the final error. Effects from such outliers can be mitigated by excluding them in the computation of the scaling factor. With respect to measures such as sMAE or sMSE, anomalous time steps/short periods of time in the series can affect the scale significantly and consequently lessen or increase the impact from such errors in the horizon unnecessarily for that particular series. However, similar to WAPE, this can be avoided by ignoring the outlier time steps from the series in the scale computation. On the other hand, when the scale is computed globally by considering all the series in the dataset, effects from such outliers in particular points of the series are mostly mitigated.

\begin{figure*}[htbp!]
	\includegraphics[scale=0.93]{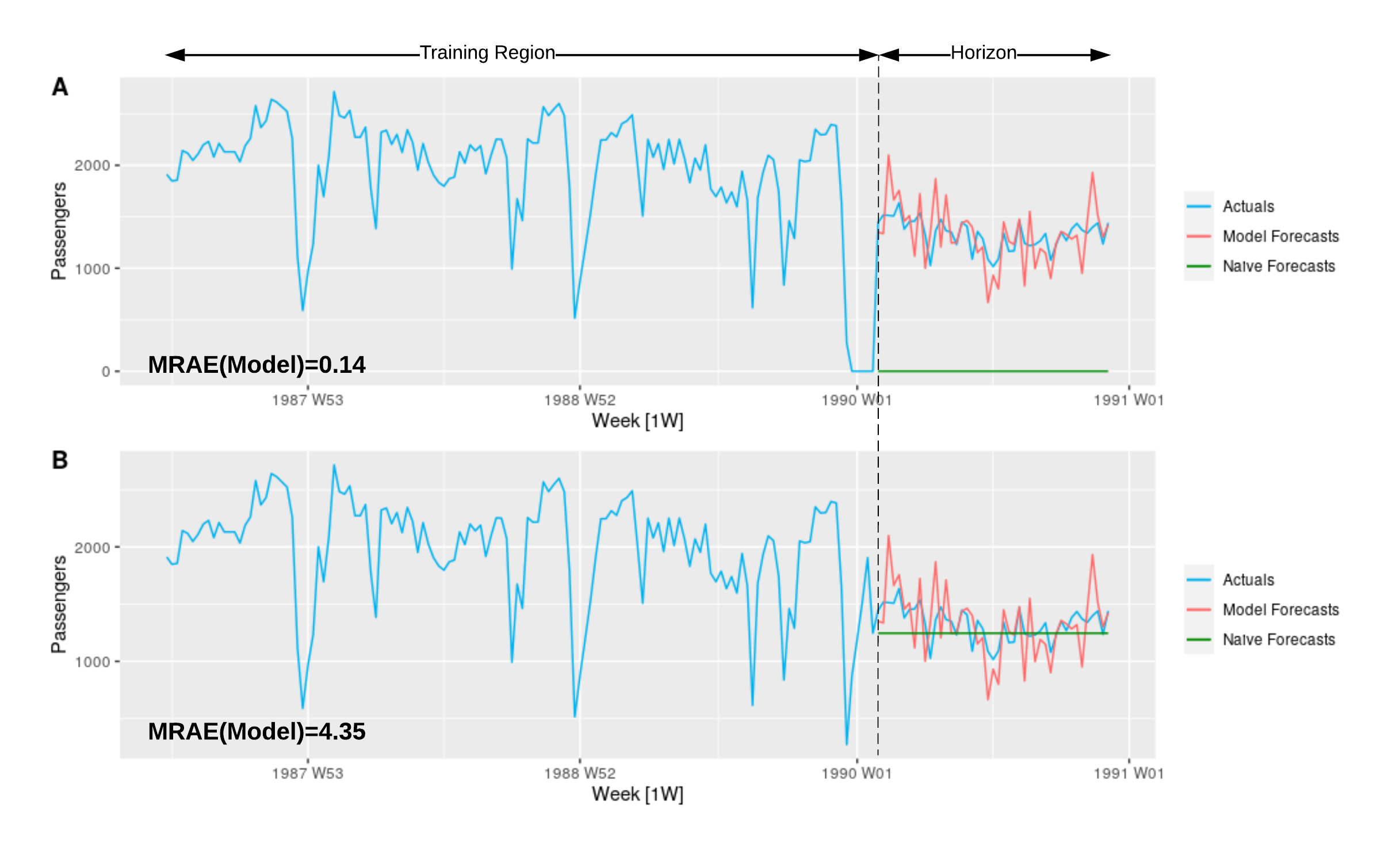}
	\caption{Relative Errors Issue - On a Series having an Outlier at the Forecast Origin}
	\label{fig:relative_errors_issues_1}
\end{figure*}

In terms of relative errors, MdRAE and GMRAE are relatively more robust to outliers, due to the used summarisation operators. Nevertheless, when computing the geometric mean, it needs to be ensured that 0 errors for both the benchmark as well as the evaluated forecasting technique are excluded. Also, MRAE is more robust with outliers than RMRSE, due to considering absolute errors in the former. However, if using the fixed origin na\"ive method as the benchmark for relative errors, its value in the forecast horizon can get affected by outliers in the last part of the training region of the series. This effectively results in large benchmark errors in the forecast horizon. This scenario is illustrated in Figure \ref{fig:relative_errors_issues_1}, where except for the outlier around the forecast origin in series $A$, both series $A$ and $B$ are equivalent in values. The model in series $A$ is robust to outliers and thus, despite the outlier in series $A$, the model produces identical forecasts for both series $A$ and $B$. On the other hand, the OOS na\"ive forecast on series $A$ is completely affected by this outlier. Due to this reason, the MRAE for the model on series $A$ is much smaller than in series $B$, although the model demonstrates the same discrepancy from the actuals in the forecast horizon on both the series. This difference of MRAE is merely due to the susceptibility of the na\"ive forecast to the outlier and not due to any misbehaviour of the model on series $B$. With Relative Measures too, the same factors as with relative errors hold on outlier series. In between the two scaled errors MASE and RMSSE, MASE is preferred to be robust over RMSSE on outliers.

\begin{figure*}[htbp!]
	\hspace{-0.5cm}
	\includegraphics[scale=1.5]{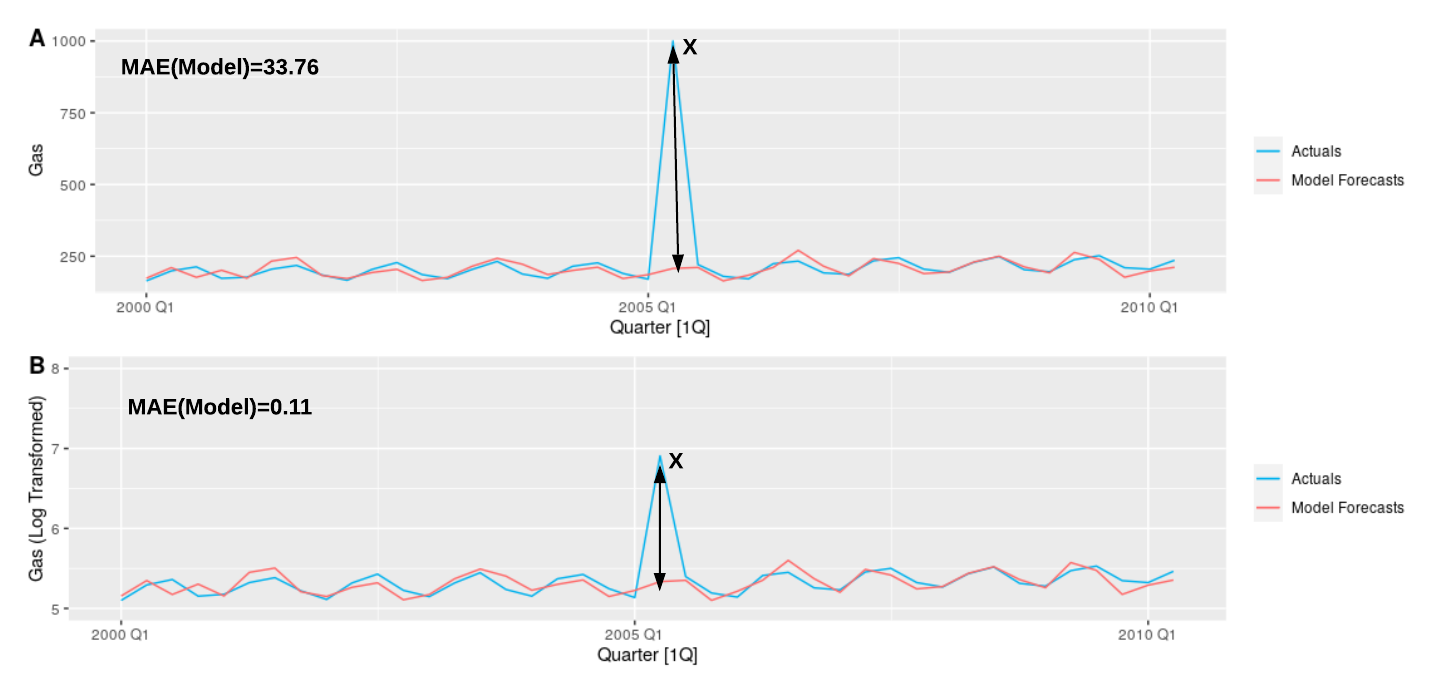}
	\caption{Transformation based Measures Issue - On a Series having an Outlier}
	\label{fig:log_errors_1}
\end{figure*}

Measures based on a transformation such as logarithm are good options to de-emphasise effects from errors at extremely large outlier points in the series. This is shown in Figure \ref{fig:log_errors_1}, where at point $X$ lies an outlier.

\section{Statistical Tests for Significance}
\label{sec:statistical_tests}


While forecast evaluation measures are critical to see the relative performance of the methods and select the best ones from their rankings, they do not give information regarding the statistical significance of the differences between these methods; i.e.\@ whether better performance of the best method is just by chance on this sample of the series or whether it is likely to dominate all the methods significantly in other samples of the data. This means that the information provided by the error metrics alone is not enough to conclude that the selected best method is the only one that should be always selected, or if there are other methods that are not significantly different from the best so that they can be used interchangeably due to their other preferable properties such as simplicity, computational efficiency etc. 

There are many ways of performing statistical significance tests reported in the literature. The general principal behind all these tests is hypothesis testing, where the null hypothesis stands for non-significance of difference in the results, and the alternative hypothesis stands for their significance. The null hypothesis is then rejected based on the value of a certain test statistic. The Diebold-Mariano test~\citep{diebold_mariano} and the Wilcoxon rank-sum test~\citep{wilcoxon} are both designed for comparing only between two competing forecasts, not necessarily methods or models. However, the Diebold-Mariano test is designed specifically for time series and parametric, meaning that it has the assumption of normality of the data whereas the Wilcoxon test is a generic non-parametric test based on the ranks of the methods. Due to considering ranks of methods for each series separately, the error measures used do not necessarily have to be scale-free. The Giacomini-White test~\citep{giacomini2006tests} again is based on the comparison of two forecasts, with the potential to assess the conditional predictive ability (CPA), a concept that refers to conditioning the choice of a potential future state of the economy, an important concept for marco economic forecasting of a small number of series. A continuation in this line of research is work by \cite{li2022conditional} that focuses on conditional superior predictive ability, in regards to a benchmark method and time series with general serial dependence. It should be noted that many of the mentioned comparison tests are per-se designed for comparing two forecasts, and a multiple testing of more than two requires a correction for multiple hypothesis testing, such as, e.g., a Bonferroni correction.

There are other techniques developed to perform comparison within a group of methods (more than 2) as well. Means of error distributions from different methods can be used to compare the mean performance of the methods. The F-test and the t-test are statistical tests in this respect. They both have parametric assumptions for the means of the error distributions, that they need to follow a normal distribution. Nevertheless, according to the Central Limit Theorem, for a sufficiently large random sample (of size $n\geq30$) from the original population, the distribution of the sample means follow an approximately normal distribution, irrespective of the distribution of the original population of errors. However, this only holds for measures such as MSE, MAE etc.\ and does not hold for e.g., RMSE, since the root of a normally distributed variable is following a chi-square distribution, which is close to normality but not equivalent. On the other hand, the Friedman test~\citep{friedman_1, friedman_2, friedman_3} is a non-parametric statistical test that can be used to detect significance between multiple competing methods, using the ranks of the methods according to mean errors. Ranks of means are equivalent to the median of the distribution~\citep{SvetunkovAdam}.

\begin{figure*}[htbp!]
	\centering
	\includegraphics[scale=0.7]{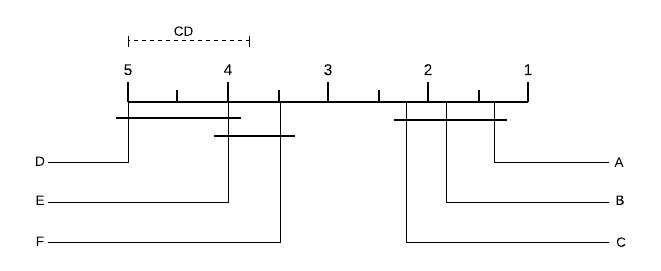}
	\caption{An example of a CD diagram to visualise the significance of the differences between a number of competing methods. The best three methods A, B and C are not significantly different from each other. On the other hand, methods D, E and F are significantly worse than those three methods. Again, the amount of data has not been enough to check whether method E is significantly better than method D or worse than method F.}
	\label{fig:cd_diagram}
\end{figure*}

However, the Friedman test only gives information on the existence of the significance between methods, but does not indicate which methods are significantly different from each other. Hence, the Friedman test is usually followed by a post-hoc test, when the null hypothesis which states that ``there are no significant differences between the methods", is rejected. There are different types of post-hoc tests, for example, the Hochberg procedure~\citep{hochberg}, the Holm process~\citep{holm}, the Bonferroni-Dunn procedure~\citep{bonferroni}, the Nemenyi method~\citep{Nemenyi1963-kh}, the Multiple Comparisons with the Best (MCB) method (practically equivalent to the Nemenyi method) or the Multiple Comparisons with the Mean (ANOM) method~\citep{anom}, and others. In general, the ANOM test holds less value in practice since it is more useful to find which methods are not significantly different from the best, than from some averagely performing method overall. The Nemenyi method works by defining confidence bounds, in terms of a Critical Distance (CD) around the mean ranks of the methods to identify which methods have overlapping confidence bounds and which do not. This method also has the added advantage that it allows to identify significant differences between two different groups of methods. As \citet{demsar} suggests, if all the comparisons are to be performed against one control method as opposed to each method against each other, procedures such as Bonferroni-Dunn and Hochberg's are better over the Nemenyi test. Once, the quantitative results for the significance of the differences are obtained using any of the aforementioned methods, they can be visualised using CD diagrams~\citep{demsar}. They are illustrated differently for the different post-hoc tests. In general, in these diagrams, a horizontal axis reports the average ranks of all the methods, and groups of methods that are not significantly different from each other are connected using black bars. This is illustrated in Figure \ref{fig:cd_diagram}, an example CD diagram.

In general, when performing significance testing, depending on the amount of data that we have, ranking can be performed either for each step in the horizon, each time series or each overall dataset. More importantly, the amount of data included heavily impacts the results of the significance tests. For example, with a very high number of series, the CD is usually very low, producing significant results for even small differences between models. This does not indicate a problem with the procedure, rather it means that the results are more reliable, that even the slightest differences between models encountered for such a large amount of data are statistically highly significant. On the other hand, it also depends on the number and the relative performance of the set of models included in the comparison. For example, having more and more poorly performing methods in the group may tend towards making the CD larger, thus making other intermediate methods have no significant difference from the best. In essence, with respect to statistical testing for significance of differences, it
is important to include a reasonable amount of data as well as a reasonable number of models with sufficient diversity to avoid spurious conclusions regarding statistical significance.

\section{Conclusions}
\label{sec:conclusions}


Model evaluation, just as in any other domain, is a crucial step in forecasting. In other major fields such as regression, classification, there exist established techniques that are the standard best practices. On the contrary, in the domain of forecasting, evaluation remains a much more complex task. 
The numerous research that has been conducted in this space over the years has contributed to many new ideas and concepts. The general trend has been to propose new methodologies to address pitfalls associated with the previously introduced. Nevertheless, for example with the forecast evaluation measures, to the best of our knowledge, all the introduced measures thus far, can break under given certain characteristics/non-stationarities of the time series. Due to the self-supervised nature of the forecasting problem, data leakage dangers need to be especially kept in mind. Furthermore, general ML practitioners and Data Scientists new to the field of forecasting are often not aware of these issues. The huge collection of forecast evaluation techniques that different practitioners use with various intentions (which are not explicitly stated), adds up to further confusion. All of this is a consequence of the lack of established best practices and guidelines for the different steps of the forecast evaluation process. Therefore, to support the ML community in this aspect, in this article we provide a compilation of common pitfalls and best practice guidelines related to forecast evaluation including data partitioning, calculating error measures etc. The key set of guidelines that we have developed are as follows.

\begin{itemize}
	\item It is always important to compare models against the right and the simplest benchmarks such as the na\"ive and the seasonal na\"ive.
	\item Using forecast plots can be misleading; making decisions purely based on the visual appeal on forecast plots is not advisable. The benchmarks and the error measures used are more important.
	\item Data leakage needs to be avoided explicitly in rolling origin evaluation and other data pre-processing tasks such as smoothing, decomposition and normalisation of the series.
	\item k-fold CV is a valid and a data efficient strategy of data partitioning for forecast model validation with pure AR based setups, when the models do not underfit the data (which can be detected with a test for serial correlation in the residuals, such as the Ljung-Box test). As such, we advise this procedure especially for short series where tsCV leads to test sets that are too small. However, if the models underfit, it is advisable to improve the models first before using any CV technique.
	\item If enough data are available, tsCV is the procedure of choice. Also, for models with a continuous state such as RNNs and ETS where the temporal order of the data is important, tsCV may be the only applicable validation strategy.
	\item There is no single globally valid evaluation measure for all scenarios. It depends on the characteristics of the data. Table \ref{tab:checklist_error_selection} provides a guide on how to select error measures to suit the data characteristics.
	\item It is advisable to evaluate the same forecasts in terms of several evaluation measures as means of sanity checking, but be aware that the forecasts cannot be optimised towards all measures together, and one particular measure will need to be chosen as the main measure. 
	\item When using statistical testing for significance of the differences between models, balancing the diversity of the compared models and against the number of data points is important to avoid spurious statistical similarity/difference between models.
\end{itemize}

Techniques available in the literature for capturing non-stationarities, such as moving average smoothing, STL decomposition etc.\@ have also been highlighted in our work. While the literature on evaluation measures is quite extensive, the exact errors (squared/absolute), summarisation operators (mean/median/geometric mean), type of scaling to use (global/per-series/per-step/, in-sample/OOS, relative/percentage) differ based on the user expectations, business utility and the characteristics of the underlying time series. Also, in real-world applications, with meaningful scales on the series, scale-dependent measures hold much value; however, the relevant scales need to be computed with respect to the underlying business utility (price/volume). 

Throughout this work, we hope to establish the knowledge and formalised guidelines related to forecast evaluation within the ML community. Due to the lack of proper knowledge in this area, ML research in the literature thus far has often either struggled to demonstrate the competitiveness of its models or arrived at spurious conclusions. It is our objective that this effort supports better and correct forecast evaluation practices within the ML community. The adoption of correct principles in this domain will certainly contribute towards developing even further superior ML based systems for forecasting. As a potential avenue for further work, it would be useful to design combination based evaluation measures for forecasting, similar to the Huber loss for model training, which is a combination of the MAE and the RMSE. These types of measures can be quite robust, combining the strengths of both measures while minimising the potential disadvantages associated with the individual measures.

\section*{Acknowledgement}
This work was done as part of the PhD degree of Hansika Hewamalage at the Faculty of IT, Monash University. This research was supported by the Australian Research Council under grant DE190100045, a Facebook Statistics for Improving Insights and Decisions research award and Monash University Graduate Research funding.

\bibliographystyle{elsarticle-harv}
\bibliography{references}
\end{document}